\documentclass[10pt,twocolumn,letterpaper]{article}

\usepackage{iccv}
\usepackage{times}
\usepackage{epsfig}
\usepackage{graphicx}
\usepackage{amsmath}
\usepackage{amssymb}
\usepackage{booktabs}
\usepackage{pifont}
\usepackage{xcolor}
\usepackage[numbers,sort]{natbib}
\newcommand{\cmark}{\textcolor{green!80!black}{\ding{51}}}
\newcommand{\xmark}{\textcolor{red}{\ding{55}}}

\newcommand{\datasetname}{TAP-Vid}

\usepackage[breaklinks=true,bookmarks=false]{hyperref} %
\iccvfinalcopy %

\ificcvfinal\fi

\begin{document}

\title{TAPIR: Tracking Any Point with per-frame Initialization and temporal Refinement}

\author{
   \bf Carl Doersch$^{*}$ \qquad Yi Yang$^{*}$ \qquad Mel Vecerik$^{*\dagger}$ \qquad Dilara Gokay$^{*}$ \qquad Ankush Gupta$^{*}$ \\
   \bf Yusuf Aytar$^{*}$ \qquad Joao Carreira$^{*}$ \qquad Andrew Zisserman$^{*\ddagger}$ \\
    $^{*}$Google DeepMind \qquad $^{\dagger}$ University College London \\
    $^{\ddagger}$VGG, Department of Engineering Science, University of Oxford
}

\maketitle
\ificcvfinal\thispagestyle{empty}\fi

\begin{abstract}
We present a novel model for Tracking Any Point (TAP) that effectively tracks any queried point on any physical surface throughout a video sequence. Our approach employs two stages: (1) a matching stage, which independently locates a suitable candidate point match for the query point on every other frame, and (2) a refinement stage, which updates both the trajectory and query features based on local correlations. The resulting model surpasses all baseline methods by a significant margin on the TAP-Vid benchmark, as demonstrated by an approximate 20\% absolute average Jaccard (AJ) improvement on DAVIS. Our model facilitates fast inference on long and high-resolution video sequences.  On a modern GPU, our implementation has the capacity to track points faster than real-time, and can be flexibly extended to higher-resolution videos.
Given the high-quality trajectories extracted from a large dataset, we demonstrate a proof-of-concept diffusion model which generates trajectories from static images, enabling plausible animations.
Visualizations, source code, and pretrained models can be found at \url{https://deepmind-tapir.github.io}. %
\end{abstract}

\section{Introduction}

The problem of point level correspondence ---i.e., determining whether two pixels in two different images are projections of the same point on the same physical surface---has long been a fundamental challenge in computer vision, with enormous potential for providing insights about physical properties and 3D shape. We consider its formulation as  ``Tracking Any Point'' (TAP)~\cite{doersch2022tap}: given a video and (potentially dense) query points on solid surfaces, an algorithm should reliably output the locations those points correspond to in every other frame where they are visible, and indicate frames where they are not -- see Fig.~\ref{fig:compare} for illustration.

\begin{figure}[t]
\begin{center}
\includegraphics[width=1.0\linewidth]{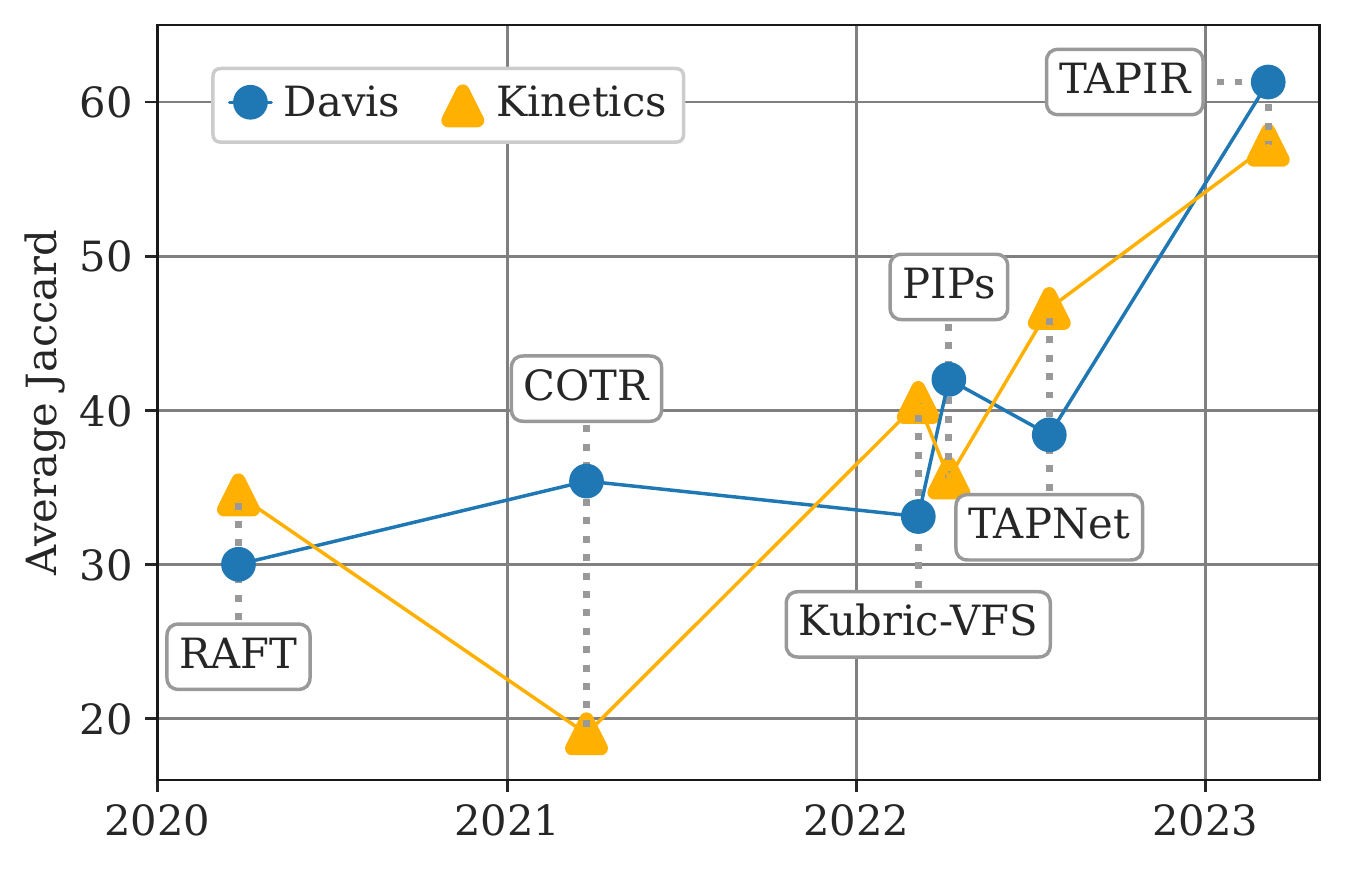}
\end{center}
   \caption{Retrospective evolution of point tracking performance over time on the recent TAP-Vid-Kinetics and TAP-Vid-DAVIS benchmarks, as measured by Average Jaccard (higher is better). In this paper we introduce TAPIR, which significantly improves performance over the state-of-the-art. This unlocks new capabilities, which we demonstrate on motion-based future prediction.}
\label{fig:fig1}
\end{figure}

Our main contribution is a new model: TAP with per-frame Initialization and temporal Refinement (TAPIR), which greatly improves over the state-of-the-art on the recently-proposed TAP-Vid benchmark~\cite{doersch2022tap}.  There are many challenges to TAP: we must robustly estimate occlusions and recover when points reappear (unlike optical flow and structure-from-motion keypoints), meaning that \emph{search} must be incorporated; yet when points remain visible for many consecutive frames, it is important to integrate information about appearance and motion across many of those frames in order to optimally predict positions.
Furthermore, little real-world ground truth is available to learn from, so supervised-learning algorithms need to learn from synthetic data without overfitting to the data distribution (i.e., sim2real).   

There are three core design decisions that define TAPIR.  The first is to use a coarse-to-fine approach. This approach has been used across many high-precision estimation problems in computer vision~\cite{lipson2022coupled, carreira2016human, yin2019hierarchical, newcombe2011kinectfusion, zollhofer2014real, wang2019prnet, su2022zebrapose, lee2017desire, marchetti2020mantra}. For our application, 
the initial `coarse' tracking consists of an occlusion-robust matching performed separately on every frame, where tracks are hypothesized using low-resolution features, without enforcing temporal continuity.   The `fine' refinement iteratively  uses local, spatio-temporal information at a higher resolution, wherein a neural network can trade-off smoothness of motion with local appearance cues to produce the most likely track.  The second design decision is to be fully-convolutional in time: the layers of our neural network consist principally of feature comparisons, spatial convolutions, and temporal convolutions, resulting in a model which efficiently maps onto modern GPU and TPU hardware.  The third design decision is that the model should estimate its own uncertainty with regard to its position estimate, in a self-supervised manner.  This ensures that low-confidence predictions can be suppressed, which improves benchmark scores.  We hypothesize that this may help downstream algorithms (e.g.\ structure-from-motion) that rely on precision, and can benefit when low-quality matches are removed.

We find that two existing architectures already have some of the pieces we need:  TAP-Net~\cite{doersch2022tap} and Persistent Independent Particles (PIPs)~\cite{harley2022particle}.  Therefore, a key contribution of our work is to effectively combine them while achieving the benefits from both.  TAP-Net  performs a global search on every frame independently, providing a coarse track that is robust to occlusions.  However, it does not make use of the continuity of videos, resulting in jittery, unrealistic tracks.  PIPs, meanwhile, gives a recipe for refinement: given an initialization, it searches over a local neighborhood and smooths the track over time.  However, PIPs processes videos sequentially in chunks, initializing each chunk with the output from the last.  The procedure struggles with occlusion and is difficult to parallelize, resulting in slow processing (i.e., 1 month to evaluate TAP-Vid-Kinetics on 1 GPU).  A key contribution of this work is observing the complementarity of these two methods.

As shown in Fig.~\ref{fig:fig1}, we find that TAPIR improves over prior works by a large margin, as measured by performance on the TAP-Vid benchmark~\cite{doersch2022tap}.   On TAP-Vid-DAVIS, TAPIR outperforms TAP-Net by $\sim$20\% while on the more challenging TAP-Vid-Kinetics, TAPIR outperforms PIPs by $\sim$20\%, and substantially reduces its inference runtime.  
To demonstrate the quality of TAPIR trajectories, we showcase a proof-of-concept model trained to \emph{generate} trajectories given individual images, and find that this model can generate plausible animations from single photographs.  

In summary, our contributions are as follows: 1) We propose a new model for long term point tracking, bridging the gap between TAP-Net and PIPs. 2) We show that the model achieves state-of-the-art results on the challenging TAP-Vid benchmark, with a significant boost on performance. 3) We provide an extensive analysis of the architectural decisions that matter for high-performance point tracking. 4) We provide a proof-of-concept of video prediction enabled by TAPIR's high-quality trajectories. Finally, 5) after analyzing components, we separately perform careful tuning of hyperparameters across entire method, in order to develop the best-performing model, which we release at \url{https://www.github.com/deepmind/tapnet} for the benefit of the community.

\section{Related Work}

Physical point correspondence and tracking have been studied in a wide range of settings. 

\noindent \textbf{Optical Flow} focuses on dense motion estimation between image pairs \cite{horn1981determining,lucas1981iterative}; methods can be divided into classical variational approaches \cite{brox2004high,brox2009large} and deep learning approaches \cite{dosovitskiy2015flownet,ilg2017flownet,ranjan2017optical,sun2018pwc,teed2020raft,xu2022gmflow}. Optical flow operates only on subsequent frames, with no simple method to track across long videos. Groundtruth is hard to obtain~\cite{butler2012naturalistic}, so it is typically benchmarked on synthetic scenes~\cite{butler2012naturalistic,dosovitskiy2015flownet,mayer2016large,sun2021autoflow}, though limited real data exists through depth scanners~\cite{geiger2012we}.

\noindent \textbf{Keypoint Correspondence} methods aim at sparse keypoint matching given image pairs, from hand-defined discriminative feature descriptors \cite{sethi1987finding,lowe1999object,lowe2004distinctive,bay2006surf} to deep learning based methods \cite{detone2018superpoint,ono2018lf,jiang2021cotr,manuelli2020keypoints}. Though it operates on image pairs, it is arguably ``long term'' as the images may be taken at different times from wide baselines. The goal, however, is typically not to track \emph{any} point, but to find easily-trackable ``keypoints'' sufficient for reconstruction. They are mainly used in structure-from-motion (SfM) settings and typically ignore occlusion, as SfM can use geometry to filter errors \cite{hartley2003multiple,schonberger2016structure,torr1999feature,vijayanarasimhan2017sfm}. 
This line of work is thus often restricted to rigid scenes \cite{zhang2000flexible,marr1979computational}, and is benchmarked accordingly~\cite{dai2017scannet,schops2017multi,balntas2017hpatches,li2018megadepth,zhou2018stereo,li2019learning}. 

\noindent \textbf{Semantic Keypoint Tracking} is often represented as keypoints or landmarks \cite{jhuang2013towards,ramanan2005strike,tompson2014real,shrivastava2017learning,xiong2013supervised,sun2014deep}. Landmarks are often ``joints'', which can have vastly different appearance depending on pose (i.e. viewed from the front or back); thus most methods rely on large supervised datasets~\cite{cheng2021equivariant,xu2021rethinking,rong2021monocular,zimmermann2019freihand,shen2015first,bartlett2006automatic}, although some works track surface meshes~\cite{von2018recovering,densepose}.  One interesting exception uses sim2real based on motion~\cite{doersch2019sim2real}, motivating better surface tracking.
Finally, some recent methods discover keypoints on moving objects~\cite{jakab2018unsupervised,zhang2018unsupervised,jakab2020self,thewlis2017unsupervised}, though these typically require in-domain training on large datasets.

\noindent \textbf{Long-term physical point tracking} is addressed in a few early works~\cite{wang2013action,sand2008particle,liu2010sift,Rubinstein2012,Sivic2006}, though such hand-engineered methods have fallen out of favor in the deep learning era. 
Our work instead builds on two recent works, TAP-Net~\cite{doersch2022tap} and Persistent Independent Particles (PIPs)~\cite{harley2022particle}, which aim to update these prior works to the deep learning era.

\section{TAPIR Model}

\begin{figure}[t]
\begin{center}
\includegraphics[width=1.0\linewidth]{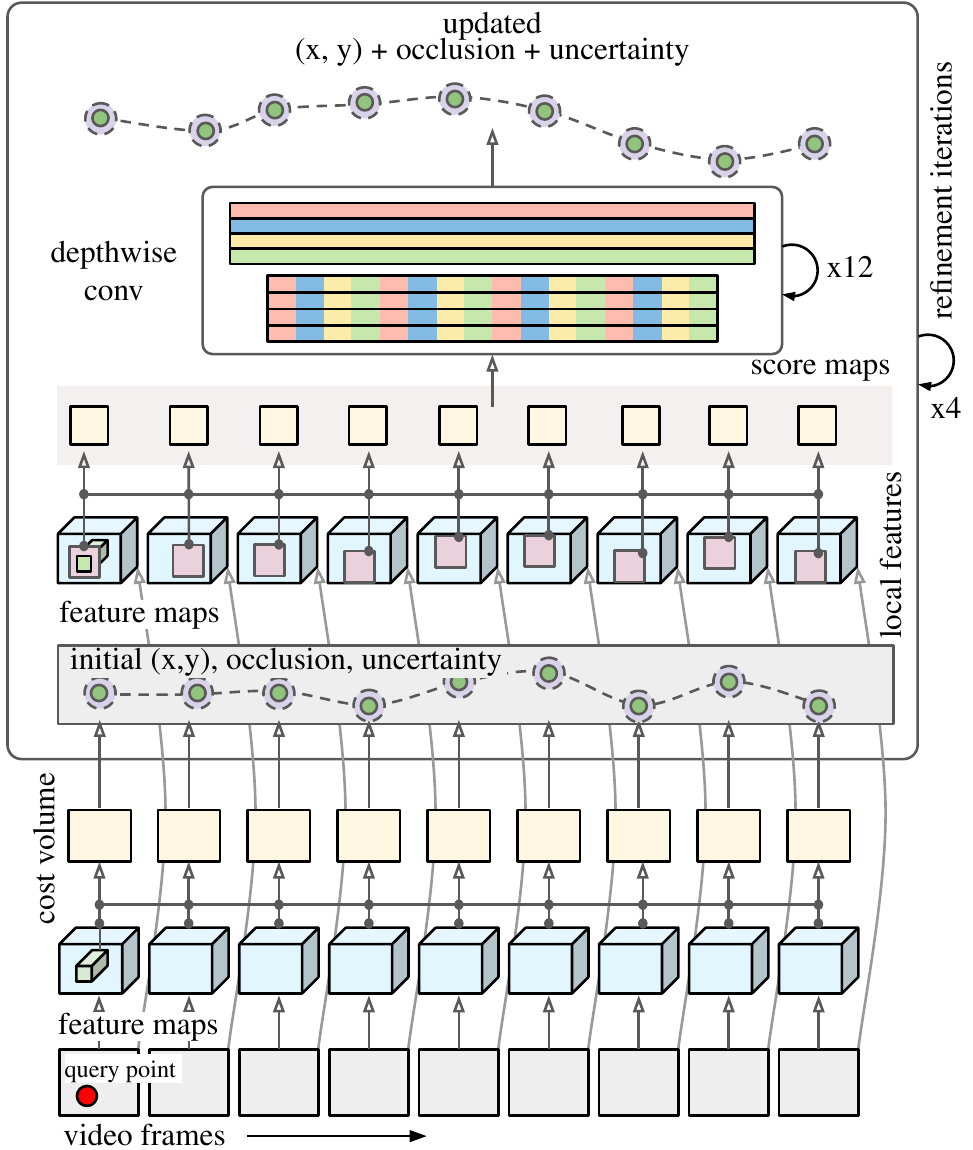}
\end{center}
   \caption{\textbf{TAPIR architecture summary.}  Our model begins with a global comparison between the query point features and the features for every other frame to compute an initial track estimate, including an uncertainty estimate.  Then, we extract features from a local neighborhood (shown in pink) around the initial estimate, and compare these to the query feature at a higher resolution, post-processing the similarities with a temporal depthwise-convolutional network to get an updated position estimate.  This updated position is fed back into the next iteration of refinement, repeated for a fixed number of iterations.  Note that for simplicity, multi-scale pyramids are not shown.}
\label{fig:arch}
\end{figure}

Given a video and a query point, our goal is to estimate the 2D location  $p_t$ that it corresponds to in every other frame $t$, as well as a 1D probability $o_t$ that the point is occluded and a 1D probability $u_t$ on the uncertainty of the estimated location.
To be robust to occlusion, we first \textit{match} candidate locations in other frames, by comparing the query features with all other features, and post-processing the similarities to arrive at an initial estimate.  Then we refine the estimate, based on the \textit{local} similarities between the query point and the target point.  Note that both stages depend mostly on {\em similarities} (dot products) between the query point features and features elsewhere (i.e.\ not on the feature alone); this ensures that we don't overfit to specific features that might only appear in the (synthetic) training data.

Fig.~\ref{fig:arch} gives an overview of our model.
We initialize an estimate of the track by finding the best match within each frame independently, ignoring the temporal nature of the video.  This involves a \emph{cost volume}: for each frame $t$, we compute the dot product between the query feature $F_{q}$ and all features in the $t$'th frame. We post-process the cost volume with a ConvNet, which produces a spatial heatmap, which is then summarized to a point estimate, in a manner broadly similar to TAP-Net.  
Given the initialization, we then compare the query feature to all features in a small region around an initial track.  We feed the comparisons into a neural network, which updates both the query features and the track.  We iteratively apply multiple such updates to arrive at a final trajectory estimate.  
TAPIR accomplishes this with a fully-convolutional architecture, operating on a pyramid of both high-dimensional, low-resolution features and low-dimensional, high-resolution features, in order to maximize efficiency on modern hardware. Finally, we add an estimate of uncertainty with respect to position throughout the architecture in order to suppress low-accuracy predictions.
The rest of this section describes TAPIR in detail.

\begin{table}
\resizebox{\columnwidth}{!}{%
\begin{tabular}{lccc}
\toprule
 & TAP-Net~\cite{doersch2022tap} & PIPs~\cite{harley2022particle} & TAPIR \\ \midrule
Per-frame Initialization & \cmark & \xmark & \cmark \\
Fully Convolutional In Time & \cmark & \xmark & \cmark \\ 
Temporal Refinement & \xmark & \cmark & \cmark \\
Multi Scale Feature & \xmark & \cmark & \cmark \\
Stride-4 Feature & \xmark & test-time only & \cmark \\
Parallel Inference & \cmark & \xmark & \cmark \\ 
Uncertainty Estimate & \xmark & \xmark & \cmark \\ 
Number of Parameters & 2.8M & 28.7M & 29.3M \\ \bottomrule
\end{tabular}%
}
\caption{Overview of models. TAPIR combines the merits from both TAP-Net and PIPs and further adds a handful of crucial components to improve performance.}
\label{tab:model_compare}
\end{table}

\subsection{Track Initialization}
The initial cost volume is computed using a relatively coarse  
feature map $F \in R^{T\times H/8\times W/8 \times C}$, where $T$ is the number of frames, $H$ and $W$ are the image height and width, and $C$ is the number of channels, computed with a standard TSM-ResNet-18~\cite{lin2020tsm} backbone.  
Features for the query point $F_q$ are extracted via bilinear interpolation at the query location, and we perform a dot product between this query feature and all other features in the video.

We compute an initial estimate of the position $p^{0}_t=(x^{0}_{t},y^{0}_{t})$ and occlusion $o^{0}_t$ by applying a small ConvNet to the cost volume corresponding to frame $t$ (which is of shape $H/8\times W/8 \times 1$ per query).  This ConvNet has two outputs: a heatmap of shape $H/8\times W/8 \times 1$ for predicting the position, and a single scalar logit for the occlusion estimate, obtained via average pooling followed by projection.  The heatmap is converted to a position estimate by a ``spatial soft argmax'', i.e., a softmax across space (to make the heatmap positive and sum to 1). Afterwards, all heatmap values which are too far from the argmax position in this heatmap are set to zero. Then, the positions for each cell of the heatmap are averaged spatially, weighted by the thresholded heatmap magnitudes.  Thus, the output is typically a location close to the heatmap's maximum; the ``soft argmax'' is differentiable, but the thresholding suppresses spurious matches, and prevents the network from ``averaging'' between multiple matches.  This can serve as a good initialization, although the model struggles to localize points to less than a few pixels' accuracy on the original resolution due to the 8-pixel stride of the heatmap.

\paragraph{Position Uncertainty Estimates.} A drawback of predicting occlusion and position independently from the cost volume (a choice that is inspired by TAP-Net)
is that if a point is visible in the ground truth, it's \emph{worse} if the algorithm predicts a vastly incorrect location than if it simply incorrectly marks the point as occluded.  After all, downstream applications may want to use the track to understand object motion; such downstream pipelines must be robust to occlusion, but may assume that the tracks are otherwise correct.  The Average Jaccard metric reflects this: predictions in the wrong location are counted as \emph{both} a ``false positive'' and a ``false negative.''  This kind of error tends to happen when the algorithm is uncertain about the position, e.g., if there are many potential matches.  Therefore, we find it's beneficial for the algorithm to also estimate its own certainty regarding position.  We quantify this by making the occlusion pathway output a second logit, estimating the probability $u^{0}_{t}$ that the prediction is likely to be far enough from the ground truth that it isn't useful, even if the model predicts that it's visible.  We define ``useful'' as whether the prediction is within a threshold $\delta$ of the ground truth.  

Therefore, our loss for the initialization for a given video frame $t$ is $\mathcal{L}(p_t^0,o_t^0,u_t^0)$, where $\mathcal{L}$ is defined as:

\begin{align}
\begin{split}
\mathcal{L}(p_t,o_t,u_t) =&  \hspace{0.2em} \mbox{Huber}(\hat{p}_t,p_t)*(1-\hat{o}_t) \\
                & +\mbox{BCE}(\hat{o}_t,o_t) \\
                & +\mbox{BCE}(\hat{u}_t,u_t)*(1-\hat{o}_t) \\
\mbox{where,}\quad \hat{u}_t =& \left\{\begin{matrix}
1 & \mbox{if } d(\hat{p}_t,p_t) > \delta \\
0 & \mbox{otherwise}
\end{matrix}\right.
\end{split}
\end{align}

Here, $\hat{o}_t\in \{0,1\}$ and $\hat{p}\in \mathbb{R}^{2}$ are the ground truth occlusion and point locations respectively, $d$ is Euclidean distance, $\delta$ is the distance threshold, Huber is a Huber loss, and BCE is binary cross entropy (both $o_t$ and $u_t$ have sigmoids applied to convert them to probabilities).  $\hat{u}_t$, the target for the uncertainty estimate $u_t$, is computed from the ground truth position and the network's prediction: it's 0 if the model's position prediction is close enough to the ground truth (within the threshold $\delta$), and 1 otherwise.  At test time, the algorithm should output that the point is visible if it's both predicted as unoccluded and if the model is confident in the prediction.  We therefore do a soft combination of the two probabilities: the algorithm outputs that the point is visible if $(1-u_t)*(1-o_t)>0.5$.

\subsection{Iterative Refinement}

Given an estimated position, occlusion, and uncertainty for each frame, the goal of each iteration $i$ of our refinement procedure is to compute an update $(\Delta p^i_t, \Delta o^i_t, \Delta u^i_t)$, which adjusts the estimate to be closer to the ground truth, integrating information across time.  The update is based on a set of ``local score maps'' which capture the query point similarity (i.e.\ dot products) to the features in the neighborhood of the trajectory.  These are computed on a pyramid of different resolutions, so for a given trajectory, they have shape $(H^{\prime}\times W^{\prime}\times L)$, where $H^{\prime}=W^{\prime}=7$, the size of the local neighborhood, and $L$ is the number of levels of the spatial pyramid (different pyramid levels are computed by spatially pooling the feature volume $F$).  As with the initialization, this set of similarities is post-processed with a network to predict the refined position, occlusion, and uncertainty estimate.  Unlike the initialization, however, we include ``local score maps'' for many frames simultaneously as input to the post-processing.  We include the current position estimate, the raw query features, and the (flattened) local score maps into a tensor of shape $T \times (C+K+4)$, where $C$ is the number of channels in the query feature, $K=H^{\prime}\cdot W^{\prime}\cdot L$ is the number of values in the flattened local score map, and 4 extra dimensions for position, occlusion, and uncertainty.  The output of this network at the $i$'th iteration is a residual $(\Delta p^i_t, \Delta o^i_t, \Delta u^i_t, \Delta F_{q,t,i})$, which is added to the position, occlusion, uncertainty estimate, and the query feature, respectively. $\Delta F_{q,t,i}$ is of shape $T\times C$; thus, after the first iteration, slightly different ``query features'' are used on each frame when computing new local score maps. 

These positions and score maps are fed into a 12-block convolutional network to compute an update $(\Delta p^i_t, \Delta o^i_t, \Delta u^i_t, \Delta F_{q,t,i})$, where each block consists of a $1\times 1$ convolution block and a depthwise convolution block.  This architecture is inspired by the
MLP-Mixer used for refinement in PIPs: we directly translate the Mixer's cross-channel layers into $1\times 1$ convolutions with the same number of channels, and the within-channel operations become similarly within-channel depthwise convolutions.  Unlike PIPs, which breaks sequences into 8-frame chunks before running the MLP-Mixer, this convolutional architecture can be run on sequences of any length.

We found that the feature maps used for computing the ``score maps'' are important for achieving high precision. The pyramid levels $l=1$ through $L-1$ are computed by spatially average-pooling the raw TSM-ResNet features $F$ at a stride of $8\cdot 2^{l-1}$, and then computing all dot products between the query feature $F_q$ and the pyramid features.  For the $t$'th frame, we extract a $7\times 7$ patch of dot products centered at $p_t$.  At training time, PIPs leaves it here, but at test time alters the backbone to run at stride 4, introducing a train/test domain gap.  We find it is effective to train on stride 4 as well, although this puts pressure on system memory.  To alleviate this, we compute a 0-th score map on the 64-channel, stride-4 input convolution map of the TSM-ResNet, and extract a comparable feature for the query from this feature map via bilinear interpolation.  Thus, the final set of local score maps has shape $(7\cdot 7 \cdot L)$. 

After $(\Delta p^i_t, \Delta o^i_t, \Delta u^i_t, \Delta F_{q,t,i})$ is obtained from the above architecture, we can iterate the refinement step as many times as desired, re-using the same network parameters.  At each iteration, we use the same loss $\mathcal{L}$ on the output, weighting each iteration the same as the initialization.

\begin{figure*}[t]
\begin{center}
\includegraphics[width=1.0\linewidth]{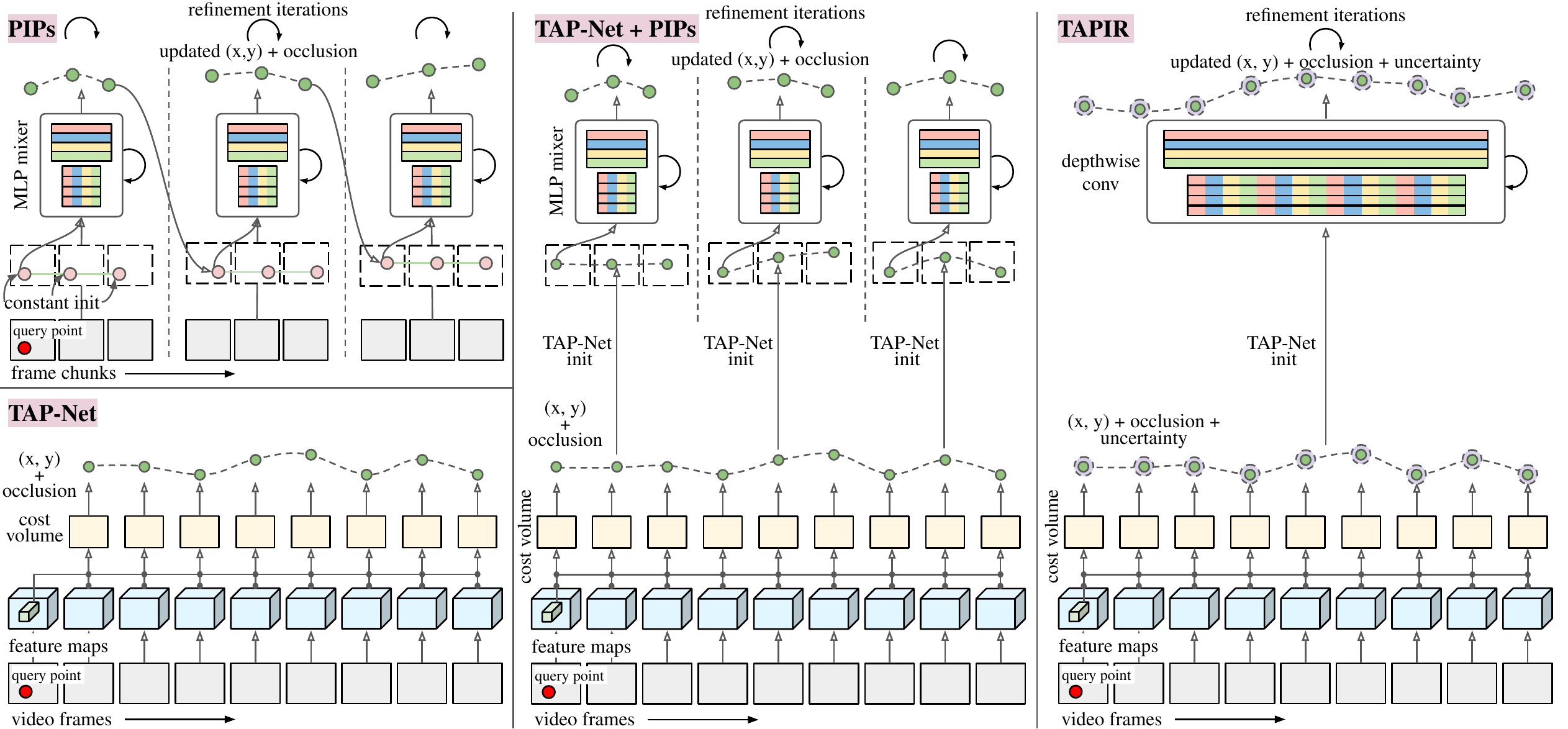}
\end{center}
   \caption{\textbf{Comparison of Architectures.} Left: the initial TAP-Net and PIPs models: TAP-Net has an independent estimate per frame, whereas PIPs breaks the video into fixed-size chunks and processes chunks sequentially.  Middle: a ``simple combination'' of these two architectures, where TAP-Net is used to initialize PIPs-style chunked iterations.  Right: Our model, TAPIR, which removes the chunking and adds uncertainty estimate. Note that for simplicity, multi-scale pyramids are not shown.}
\label{fig:arch_compare}
\end{figure*}

\noindent \textbf{Discussion} As there is some overlap with prior work, Figure~\ref{fig:arch_compare} depicts the relationship between TAPIR and other architectures.  For the sake of ablations, we also develop a ``simple combination'' model, which is intended to be a simpler combination of a TAP-Net-style initialization with PIPs-style refinements, and follows prior architectural decisions regardless of whether they are optimal.  To summarize, the major architectural decisions needed to create the ``simple combination'' of TAP-Net and PIPs (and therefore the contributions of this model) are as follows.  First we exploit the complementarity between TAP-Net and PIPs.  Second, we remove ``chaining'' from PIPs, replacing that initialization with TAP-Net's initialization.  We directly apply PIPs' MLP-Mixer architecture by `chunking' the input sequence.  Note that we adopt TAP-Net's feature network for both initialization and refinement.  Finally, we adapt the TAP-Net backbone to extract a multi-scale and high resolution pyramid (follow PIPs stride-4 test-time architecture, but matching TAP-Net's end-to-end training).  We apply losses at all layers of the hierarchy (whereas TAP-Net has a single loss, PIPs has no loss on its initialization).  The additional contributions of TAPIR are as follows: first, we entirely replace the PIPs MLP-Mixer (which must be applied to fixed-length sequences) with a novel depthwise-convolutional architecture, which has a similar number of parameters but works on sequences of any length.  Second, we introduce uncertainty estimate, which are computed at every level of the hierarchy.  These are `self-supervised' in the sense that the targets depend on the model's own output.

\subsection{Training Dataset}
Although TAPIR is designed to be robust to the sim2real gap, we can improve performance by minimizing the gap itself.  One subtle difference between the Kubric MOVi-E~\cite{kubric} dataset and the real world is the lack of \emph{panning}: although the Kubric camera moves, it is always set to ``look at'' a single point at the center of the workspace.  Thus, stationary parts of the scene rotate around the center point of the workspace on the 2D image plane.  Models thus tend to fail on real-world videos with panning.  Therefore, we modified the MOVi-E dataset so that the ``look at'' point moves along a random linear trajectory.  The start and end points of this trajectory are enforced to be less than 1.5 units off the ground and lie within a radius of 4 units from the workspace center; the trajectory must also pass within 1 unit of the workspace center at some point, to ensure that the camera is usually looking toward the objects in the scene. See Appendix~\ref{sec:training_dataset} for details.

\section{Extension to High-Resolution Videos}
Although TAPIR is trained only on $256 \times 256$ videos, it depends mostly on comparisons between features, meaning that it might extend trivially to higher-resolution videos by simply using a larger convolutional grid, similar to ConvNets.  However, we find that applying the per-frame initialization to the entire image is likely to lead to false positives, as the model is tuned to be able to discriminate between a $32\times 32$ grid of features, and may lack the specificity to deal with a far larger grid.  Therefore, we instead create an image pyramid, by resizing the image to $K$ different resolutions.  The lowest resolution is $256 \times 256$, while the highest resolution is the original resolution, with logarithmically-spaced resolutions in between that resize at most by a factor of 2 at each level.  We run TAPIR to get an initial estimate at $256\times 256$, and then repeat iterative refinement at every level, with the same number of iterations at every level.  When running at a given level, we use the position estimate from the prior level, but we directly re-initialize the occlusion and uncertainty estimates from the track initialization, as we find the model otherwise tends to become overconfident.  The final output is then the average output across all refinement resolutions.

\section{Experiments}

We evaluate on the TAP-Vid~\cite{doersch2022tap} benchmark, a recent large-scale benchmark evaluating the problem of interest.  TAP-Vid consists of point track annotations on four different datasets, each with different challenges.  DAVIS~\cite{pont20172017} is a benchmark designed for tracking, and includes complex motion and large changes in object scale. Kinetics~\cite{carreira2017quo} contains over 1000 labeled videos, and contains all the complexities of YouTube videos, including difficulties like cuts and camera shake.  RGB Stacking is a synthetic dataset of robotics videos with precise point tracks on large textureless regions.  Kubric MOVi-E, used for training, contains realistic renderings of objects falling onto a flat surface; it comes with a validation set that we use for comparisons. We conduct training exclusively on the Kubric dataset, and selected our best model primarily by observing DAVIS.  We performed no automated evaluation for hyperparameter tuning or model selection on any dataset.  We train and evaluate at a resolution of $256\times 256$.

\paragraph{Implementation Details}
We train for 50,000 steps with a cosine learning rate schedule, with a peak rate of $1\times10^{-3}$ and 1,000 warmup steps.  We use 64 TPU-v3 cores, using 4 24-frame videos per core in each step. We use an Adam-W optimizer with $\beta_1=0.9$ and $\beta_2=0.95$ and weight decay $1\times10^{-2}$. For most experiments, we use cross-replica batch normalization within the (TSM-)ResNet backbone only, although for our public model we find instance normalization is actually more effective.  For most experiments we use $L=5$ to match PIPs, but in practice we find $L>3$ has little impact (See Appendix~\ref{sec:pyramid_layers}).  Therefore, to save computation, our public model uses $L=3$. See Appendix~\ref{sec:tapir_details} for more details.

\subsection{Quantitative Comparison}

\begin{table*}[t]
\begin{center}
\resizebox{\linewidth}{!}{ %
\begin{tabular}{l|ccc|ccc|ccc|ccc}
\toprule
 & \multicolumn{3}{c|}{Kinetics} & \multicolumn{3}{c|}{Kubric} & \multicolumn{3}{c|}{DAVIS} & \multicolumn{3}{c}{RGB-Stacking} \\
Method &  AJ & $<\delta^{x}_{avg}$ & OA &  AJ & $<\delta^{x}_{avg}$ & OA &  AJ & $<\delta^{x}_{avg}$ & OA &  AJ & $<\delta^{x}_{avg}$ & OA \\
\midrule
COTR~\cite{jiang2021cotr} & 19.0  & 38.8  & 57.4  &  40.1 & 60.7 & 78.55 & 35.4 & 51.3 & 80.2 & 6.8 & 13.5 & 79.1 \\
Kubric-VFS-Like~\cite{kubric} & 40.5 & 59.0 & 80.0 & 51.9 & 69.8 & 84.6 & 33.1 & 48.5 & 79.4 & 57.9 & 72.6 & 91.9 \\
RAFT~\cite{teed2020raft} & 34.5 & 52.5 & 79.7 & 41.2 & 58.2 & 86.4 & 30.0 & 46.3 & 79.6 & 44.0 & 58.6 & 90.4 \\
PIPs~\cite{harley2022particle} & 35.3 & 54.8 & 77.4 & 59.1 & 74.8 & 88.6 & 42.0 & 59.4 & 82.1 & 37.3 & 51.0 & 91.6 \\
TAP-Net~\cite{doersch2022tap} & 46.6 & 60.9 & 85.0 & 65.4 & 77.7 & 93.0 & 38.4 & 53.1 & 82.3 & 59.9 & 72.8 & 90.4 \\
\midrule
TAPIR (MOVi-E) & 57.1 & 70.0 & 87.6 & 84.3 & 91.8 & 95.8 & 59.8 & 72.3 & 87.6 & \textbf{66.2} & \textbf{77.4} & \textbf{93.3} \\
TAPIR (Panning MOVi-E) & \textbf{57.2} & \textbf{70.1} & \textbf{87.8} & \textbf{84.7} & \textbf{92.1} & \textbf{95.8} & \textbf{61.3} & \textbf{73.6} & \textbf{88.8} & 62.7 & 74.6 & 91.6 \\
\bottomrule
\end{tabular}
}
\caption{{\bf Comparison of TAPIR to prior results on \datasetname}.  $<\delta^{x}_{avg}$ is the fraction of points unoccluded in the ground truth, where the prediction is within a threshold, ignoring the occlusion estimate; OA is the accuracy of predicting occlusion.  AJ is Average Jaccard, which considers both position and occlusion accuracy.}
\label{tab:results}
\end{center}
\end{table*}

\begin{figure*}[t]
\begin{center}
\includegraphics[width=1.0\textwidth]{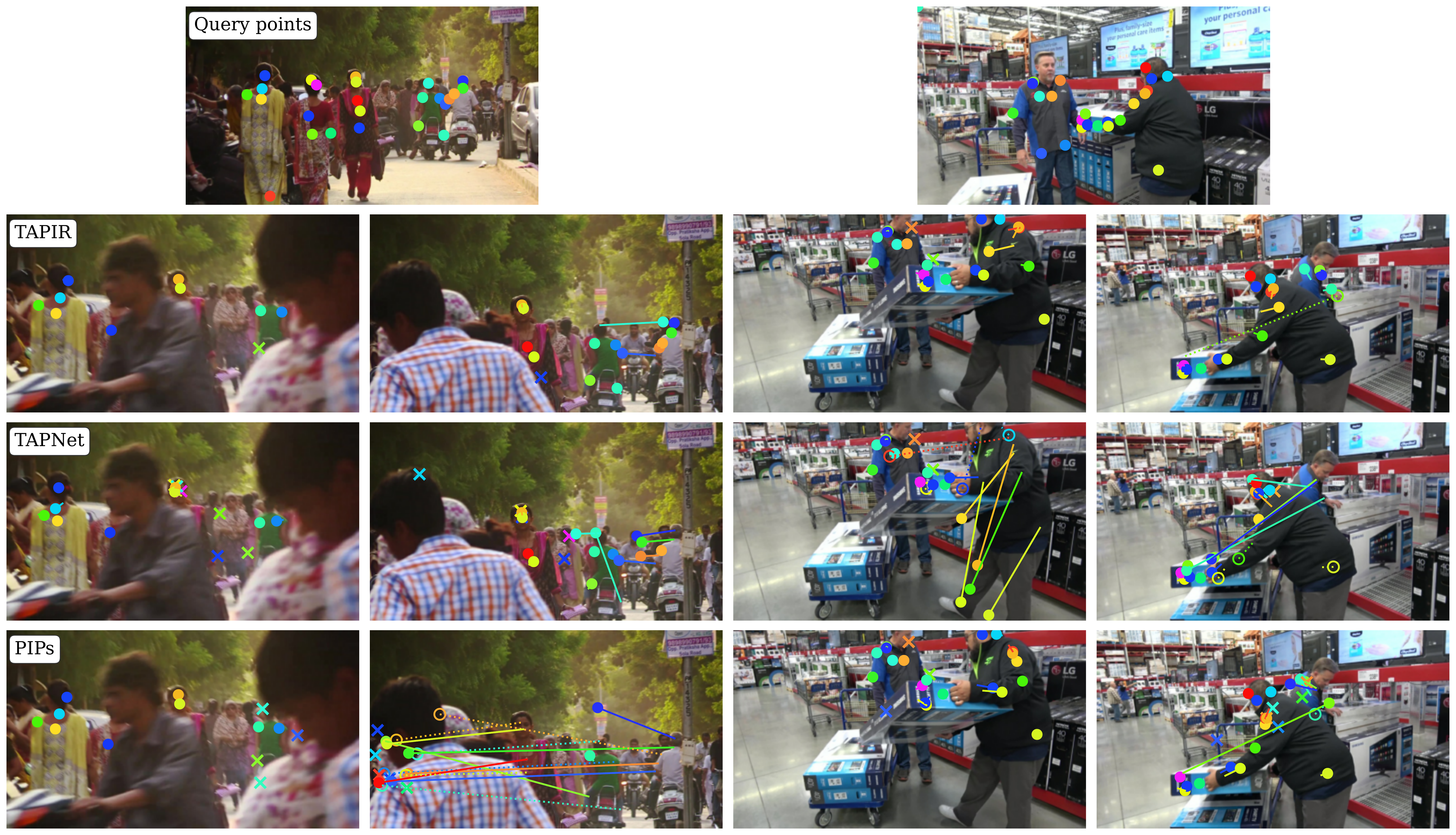}
\end{center}
\caption{\textbf{TAPIR compared to TAP-Net and PIPs.} Top shows the query points on the video's first frame.  Predictions for two later frames are below.  We show predictions (filled circles) relative to ground truth (GT) (ends of the associated segments).  x's indicate predictions where the GT is occluded, while empty circles are  points visible in GT but predicted occluded (note: models still predict position regardless of occlusion).  The majority of the street scene (left) gets occluded by a pedestrian; as a result, PIPs loses many points.  TAP-Net fails on the right video, possibly because the textureless clothing is difficult to match without relying on temporal continuity.  TAPIR, meanwhile, has far fewer failures.}
\label{fig:compare}
\end{figure*}

Table~\ref{tab:results} shows the performance of TAPIR relative to prior works.  TAPIR improves by 10.6\% absolute performance over the best model on Kinetics (TAP-Net) and by 19.3\% on DAVIS (PIPs).  These improvements are quite significant according to the TAP-Vid metrics: Average Jaccard consists of 5 thresholds, so under perfect occlusion estimation, a 20\% improvement requires improving \textit{all} points to the next distance threshold (half the distance). It's worth noting that, like PIPs, TAPIR's performance on DAVIS is slightly higher than on Kinetics, whereas for TAP-Net, Kinetics has higher performance, suggesting that TAPIR's reliance on temporal continuity isn't as useful on Kinetics videos with cuts or camera shake.  Note that the bulk of the improvement comes from the improved model rather than the improved dataset (TAPIR (MOVi-E) is trained on the same data as TAP-Net, while `Panning' is the new dataset).  The dataset's contribution is small, mostly improving DAVIS, but we found qualitatively that it makes the most substantial difference on backgrounds when cameras pan.  This is not well represented in TAP-Vid's query points, but may matter for real-world applications like stabilization.  RGB-stacking, however, is somewhat of an outlier: training on MOVi-E is still optimal (unsurprising as the cameras are static), and there the improvement is only $6.3\%$, possibly because TAPIR does less to help with textureless objects, indicating an area for future research. Fig.~\ref{fig:compare} shows some example predictions on DAVIS.  TAPIR corrects many types of errors, including problems with temporal coherence from TAP-Net and problems with occlusion from PIPs. We include further qualitative results on our project webpage, which illustrate how noticeable a ~20\% improvement is. 

\begin{table}[t]
\resizebox{\columnwidth}{!}{%
\begin{tabular}{l|ccc|ccc|ccc}
\toprule
 & \multicolumn{3}{c|}{Kinetics} & \multicolumn{3}{c|}{DAVIS} & \multicolumn{3}{c}{RGB-Stacking} \\ %
Method & AJ & $<\delta^{x}_{avg}$ & OA & AJ & $<\delta^{x}_{avg}$ & OA & AJ & $<\delta^{x}_{avg}$ & OA \\
\midrule
TAP-Net & 38.5 & 54.4 & 80.6 & 33.0 & 48.6 & 78.8 & 53.5 & 68.1 & 86.3 \\
TAP-Net (Panning MOVi-E) & 39.5 & 56.2 & 81.4 & 36.0 & 52.9 & 80.1 & 50.1 & 65.7 & 88.3 \\
TAPIR (MOVi-E) & 49.6 & 64.2 & \textbf{85.2} & 55.3 & 69.4 & 84.4 & \textbf{56.2} & \textbf{70.0} & \textbf{89.3} \\
TAPIR (Panning MOVi-E) & \textbf{49.6} & \textbf{64.2} & 85.0 & \textbf{56.2} & \textbf{70.0} & \textbf{86.5} & 55.5 & 69.7 & 88.0 \\
\bottomrule
\end{tabular}%
}
\caption{\textbf{Comparison under query first metrics.}  We see largely the same relative performance trend whether the model is queried in a `strided' fashion or whether the model is queried with the first frame where the point appeared, although performance is overall lower than the `strided' evaluation. }
\label{tab:query_first_compare}
\end{table}

TAP-Vid proposes another evaluation for videos at the resolution of $256 \times 256$ called `query-first.'  In the typical case (Table~\ref{tab:results}), the query points are sampled in a `strided' fashion from the ground-truth tracks: the same trajectory is therefore queried with multiple query points, once every 5 frames.  In the `query first' evaluation, only the first point along each trajectory is used as a query.  These points are slightly harder to track than those in the middle of the video, because there are more intervening frames wherein the point's appearance may change.   Table~\ref{tab:query_first_compare} shows our results on this approach.  We see that the performance broadly follows the results from the `strided' evaluation: we outperform TAP-Net by a wide margin on real data (10\% absolute on Kinetics, 20\% on DAVIS), and by a smaller margin on RGB-Stacking (3\%).  This suggests that TAPIR remains robust even if the output frames are far from the query point.

We also conduct a comparison of computational time for model inference on the DAVIS video \textit{horsejump-high} with a resolution of 256x256. All models are evaluated on a single GPU using 5 runs on average. With Just In Time (JIT) Compilation with JAX, for 50 query points randomly sampled on the first frame, TAP-Net finishes within 0.1 seconds and TAPIR finishes within 0.3 seconds, i.e., roughly 150 frames per second.  This is possible because TAPIR, like TAP-Net, maps efficiently to GPU parallel processing. In contrast, PIPs takes 34.5 seconds due to a linear increase with respect to the number of points and frames processed. See Appendix~\ref{sec:runtime} for more details. 

\subsection{High-Resolution Results}

\begin{table}[t]
\begin{center}
\resizebox{\linewidth}{!}{ %
\begin{tabular}{l|ccc|ccc}
\toprule
 & \multicolumn{3}{c|}{Kinetics} & \multicolumn{3}{c}{DAVIS} \\
Method &  AJ & $<\delta^{x}_{avg}$ & OA &  AJ & $<\delta^{x}_{avg}$ & OA  \\
\midrule
TAPIR $256\times 256$ & 57.2 & 70.1 & 87.8 & 61.3 & 73.6 & 88.8 \\
TAPIR Hi-Res & 60.0 & 72.1 & 86.7 & 65.7 & 77.6 & 86.7 \\
\bottomrule
\end{tabular}
}
\caption{{\bf TAPIR at high resolution on \datasetname}.  Each video is resized so it is at most $1080$ pixels tall and $1920$ pixels wide for DAVIS, and $720$ pixels tall and $1280$ pixels wide for Kinetics.}
\label{tab:hires}
\vspace{-0.3cm}
\end{center}
\end{table}

For running TAPIR on high-resolution videos, we partitioned the model across 8 TPU-v3 devices, where each device received a subset of frames.  This allows us to fit DAVIS videos at 1080p and Kinetics at 720p without careful optimization of JAX code.  Results are shown in Table~\ref{tab:hires}.
Unsurprisingly, having more detail in the image helps the model localize the points better.  However, we see that occlusion prediction becomes less accurate, possibly because large context helps with occlusion.  Properly combining multiple resolutions is an interesting area for future works.

\subsection{Ablation Studies}

\paragraph{TAPIR vs Simple Combination} We analyze the contributions of the main ideas.  Our first contribution is to combine PIPs and TAP-Net (Fig.~\ref{fig:arch_compare}, center). Comparing this directly to PIPs is somewhat misleading:  running the open-source `chaining' code on Kinetics requires roughly 1 month on a V100 GPU (roughly 30 minutes to process a 10-second clip).  But how crucial is chaining?  After all, PIPs MLP-mixers have large spatial receptive fields due to their multi-resolution pyramid.  Thus, we trained a na\"ive version of PIPs without chaining: it initializes the entire track to be the query point, and then processes each chunk independently.  For a fair comparison, we train using the same dataset and feature extractor, and other hyperparameters as TAPIR, but retain the 8-frame MLP Mixer with the same number of iterations as TAPIR.  Table~\ref{tab:model_ablation} shows that this algorithm (Chain-free PIPs) is surprisingly effective, on par with TAP-Net. Next, we combine the Chain-free PIPs with TAP-Net, replacing the constant initialization with TAP-Net's prediction. The ``simple combination'' (Fig.~\ref{fig:arch_compare}, center) alone improves over PIPs, Chain-free PIPs, and TAP-Net (i.e. 43.3\% v.s. 50.0\% on DAVIS).  However, the other new ideas in TAPIR also account for a substantial portion of the performance (i.e. 50.0\% v.s. 61.3\% on DAVIS).

\begin{table}[t]
\resizebox{\columnwidth}{!}{%
\begin{tabular}{lcccc}
\toprule
Average Jaccard (AJ) & Kinetics & DAVIS & RGB-Stacking & Kubric \\ \midrule
PIPs~\cite{harley2022particle} & 35.3 & 42.0 & 37.3 & 59.1 \\
TAP-Net~\cite{doersch2022tap} & 48.3 & 41.1 & 59.2 & 65.4 \\
Chain-Free PIPs & 47.2 & 43.3 & 61.3 & 75.3 \\
Simple Combination & 52.9 & 50.0 & 62.7 & 78.1 \\
TAPIR & \textbf{57.2} & \textbf{61.3} & \textbf{62.7} & \textbf{84.7} \\ \bottomrule
\end{tabular}
}
\caption{\textbf{PIPs, TAP-Net, and a simple combination vs TAPIR.} We compare TAPIR (Fig. \ref{fig:arch_compare} right) to TAP-Net, PIPs (Fig. \ref{fig:arch_compare} left), and the ``simple combination'' (Fig. \ref{fig:arch_compare} center).  Chain-Free PIPs is our reimplementation of PIPs that removes the extremely slow chaining operation and uses the TAPIR backbone network.  All models in this table are trained on Panning MOVi-E except PIPs, which uses its own dataset.}
\label{tab:model_ablation}
\end{table}

\begin{table}[t]
\resizebox{\columnwidth}{!}{%
\begin{tabular}{lcccc}
\toprule
Average Jaccard (AJ) & Kinetics & DAVIS & RGB-Stacking & Kubric \\ \midrule
Full Model & \textbf{57.2} & \textbf{61.3} & \textbf{62.7} & \textbf{84.7} \\
- No Depthwise Conv & 54.9 & 53.8 & 61.9 & 79.7 \\
- No Uncertainty Estimate & 54.4 & 58.6 & 61.5 & 83.4 \\
- No Higher Res Feature & 54.0 & 54.0 & 59.5 & 80.4 \\
- No TAP-Net Initialization & 54.7 & 59.3 & 60.3 & 84.1 \\
- No Iterative Refinement & 48.1 & 41.6 & 60.3 & 64.9 \\
 \bottomrule
\end{tabular}%
}
\caption{\textbf{Model ablation by removing one major component at a time.} -No Depthwise Conv uses a MLP-Mixer on a chunked video similar to PIPs. -No Uncertainty Estimate uses the losses directly as described in TAP-Net. -No Higher Res Feature uses only 4 pyramid levels in iterative refinement and has no features at stride 4. -No TAP-Net Initialization uses a constant initialization for iterative refinement. All the components contribute non-trivially to the performance of the model.}
\label{tab:component_ablation}
\end{table}

\paragraph{Model Components} Table~\ref{tab:component_ablation} shows the effect of our architectural decisions. First, we consider two novel architectural elements: the depthwise convolution which eliminates chunking from the PIPs model, and the uncertainty estimate. We see for Kinetics, both methods contribute roughly equally to performance, while for DAVIS, the depthwise conv is more important.  A possible interpretation is that DAVIS contains more temporal continuity than Kinetics, so chunking is more harmful.  
Higher-resolution features are also predictably important to performance, especially on DAVIS.  Table~\ref{tab:component_ablation} also considers a few other model components that are not unique to TAPIR.  TAP-Net initialization is important, but the model still gives reasonable performance without it.  Interestingly, the full model without refinement (i.e. TAP-Net trained on Panning MOVi-E with the uncertainty estimate) performs similarly to TAP-Net trained on Panning MOVi-E (Table~\ref{tab:model_ablation}, second line); one possible interpretation is that TAP-Net is overfitting to its training data, leading to biased uncertainty estimate.

\begin{table}[t]
\resizebox{\columnwidth}{!}{%
\begin{tabular}{lcccc}
\toprule
Average Jaccard (AJ) & Kinetics & DAVIS & RGB-Stacking & Kubric \\ \midrule
0 iterations & 48.1 & 41.6 & 60.3 & 64.9 \\
1 iteration & 55.7 & 55.0 & 62.7 & 80.1 \\
2 iterations & 56.7 & 58.8 & 62.0 & 83.1 \\
3 iterations & 56.8 & 60.6 & 60.9 & 84.1 \\
4 iterations & 57.2 & 61.3 & \textbf{62.7} & \textbf{84.7} \\
5 iterations & \textbf{57.8} & \textbf{61.6} & 61.5 & 84.3 \\
6 iterations & 57.2 & 60.4 & 56.6 & 82.9 \\ \bottomrule
\end{tabular}%
}
\caption{\textbf{Evaluation performance against the number of iterative updates.} It can be observed that after 4 iterations, the performance no longer improves significantly. This could be attributed to the fact that TAP-Net already provides a good initialization.}
\label{tab:num_iteration}
\end{table}

\paragraph{Iterations of Refinement} Finally, Table~\ref{tab:num_iteration} shows how performance depends on the number of refinement iterations. We use the same number of iterations during training and testing for all models. Interestingly, we observed the best performance for TAPIR at 4-5 iterations, whereas PIPs used 6 iterations. Likely a stronger initialization (i.e., TAP-Net) requires fewer iterations to converge. The decrease in performance with more iterations implies that there may be some degenerate behavior, such as oversmoothing, which may be a topic for future research. In this paper, we use 4 iterations for all other experiments.

Further model ablations, including experiments with RNNs and dense convolutions as replacements for our depthwise architecture, the number of feature pyramid layers, and the importance of time-shifting in the features, can be found in Appendix~\ref{sec:ablation}.  Depthwise convolutions work as well or better than the alternatives while being less expensive.  We find little impact of adding more feature pyramid layers as PIPs does, so we remove them from TAPIR.  Finally, we find the impact of time-shifting to be minor.

\section{Open-Source Version}

In the previous sections, our goal was to align our decisions with those made by prior models TAP-Net and PIPs, allowing for easier comparisons and identification of crucial high-level architectural choices. However, in the interest of open-sourcing the most powerful model possible to the community, we now conduct more detailed hyperparameter tuning. This comprehensive model, inclusive of training and inference code, is openly accessible at \href{https://github.com/deepmind/tapnet}{https://github.com/deepmind/tapnet}. The open-source TAPIR model introduces some extra model modifications including (1) the backbone, (2) several constants utilized in the model and training process, and (3) the training setup.

Regarding the backbone, we employ a ResNet-style architecture based on a single-frame design, structurally resembling the TSM-ResNet used in other instances but with an additional ResNet layer. Specifically, we utilize a Pre-ResNet18 backbone with the removal of the max-pooling layer. The four ResNet layers consist of feature dimensions of [64, 128, 256, 256], with 2 ResNet blocks per layer. The convolution strides are set as [1, 2, 2, 1]. Apart from the change in the backbone, the TAP features in the pyramid originate from ResLayer2 and ResLayer4, corresponding to resolutions of 64×64 and 32×32, respectively, on a 256×256 image. It is noteworthy that the dimensionality of high-resolution feature map has been altered from 64 to 128. Another significant modification involves transitioning from batch normalization to instance normalization. Our experiments reveal that instance normalization, in conjunction with the absence of a max-pooling layer, yields optimal results across the TAP-Vid benchmark datasets. Collectively, these alterations result in an overall improvement of 1\% on DAVIS and Kinetics datasets.

Regarding the constants, we have discovered that employing a softmax temperature of 20.0 performs marginally better than 10.0. Additionally, we have chosen to use only 3 pyramid layers instead of the 5 employed elsewhere. Consequently, a single pyramid layer consists of averaging the ResNet output. For our publicly released version, we have opted for 4 iterative refinements. Furthermore, the expected distance threshold has been set to 6.

In terms of optimization, we train the model for 50,000 steps, which we have determined to be sufficient for convergence to a robust model. Each TPU device is assigned a batch size of 8.  We use 1000 warm-up steps and a base learning rate of 0.001. To manage the learning rate, we employ a cosine learning rate scheduler with a weight decay of 0.1. The AdamW optimizer is utilized throughout the training process.

We train TAPIR using both our modified panning dataset and the publicly available Kubric MOVi-E dataset. Our findings indicate that the model trained on the public Kubric dataset performs well when the camera remains static. However, it may produce suboptimal results on videos with moving cameras, particularly for points that leave the image frame. Consequently, we have made both versions of the model available, allowing users to select the version that best suits their specific application requirements.

\begin{table}[t]
\resizebox{\columnwidth}{!}{%
\begin{tabular}{lcccc}
\toprule
Average Jaccard (AJ) & Kinetics & DAVIS & RGB-Stacking & Kubric \\ \midrule
TAPIR tuned (MOVi-E) & 60.2 & 62.9 & 73.3 & 88.3 \\ 
TAPIR tuned (Panning Kubric) & 58.2 & 62.4 & 69.1 & 85.8 \\ \bottomrule
\end{tabular}
}
\caption{\textbf{Open Sourced TAPIR results on TAP-Vid Benchmark.} Comparing to our major reported model, the open sourced version improves substantially, particularly on RGB-Stacking dataset.}
\label{tab:open_source_model}
\end{table}

\section{Animating still images with TAPIR}

Generative modeling of images and videos has experienced an explosion in popularity due to recent striking text-conditioned generation results using diffusion models~\cite{saharia2022photorealistic, ramesh2022hierarchical, nichol2021glide, gu2022vector, ho2022video}.  Animating still images remains extremely challenging, due to the high computational cost of video modeling and the difficulty of generating realistic motion.  Is it possible to improve the generated videos via an explicit representation of surface motion?  %
Here, we develop a proof-of-concept for a diffusion-based system that uses TAPIR tracks to do this.  

We perform video prediction using a pair of diffusion models: one which generates \emph{dense trajectories} given an input image (the ``trajectory'' model), and a second which takes the input image and the trajectories and generates pixels (the ``pixel'' model).  At test time, the outputs from the trajectory model dictate the motion that should be present in the video, ideally ensuring physical plausibility.
The pixel model uses the trajectories to attend to the appropriate location in the first frame, in order to produce appearance which is consistent across the video.  %

For the trajectory model, we encode the conditioning image with a
small ConvNet, which produces a dense grid of features at stride 4.  This is used as conditioning for a U-Net that performs diffusion denoising on the trajectories.
We find that encoding the trajectories is important: for each frame, we include the raw $x$ and $y$ positions, both in absolute coordinates and coordinates relative to the starting position, and we also include a position encoding of the absolute locations with 32 Fourier features.  The U-Net architecture includes self-attention layers, so the Fourier features allow the model to measure whether trajectories are spatially nearby late in an animation.  
For simplicity and memory efficiency, we stack these values for all frames along the channel axis: thus, the input is a 2D grid of shape $H//4 \times W//4 \times (T \cdot (1+2+2+32))$, (occlusion, relative position, absolute position, fourier encoding), which is processed with a 2D U-Net.
The model predicts the clean data given the noisy data, which is the typical ``clean data prediction'' diffusion setup.  %
The pixel model is a more traditional image generation model, which predicts the noise residual given noisy pixels.  It takes three inputs, and outputs a single denoised frame.  The first input is 3 noisy frames, allowing the model to infer some temporal context.  The second is the first frame, warped according to the trajectories from the trajectory model.  The third is features for the first frame, computed by a standard ConvNet before, being warped according to the same trajectories.  We train on an internal dataset of 1M video clips, processed into 24-frame clips at $256\times 256$ resolution.  See supplementary for more details.

\begin{figure}[t]
\begin{center}
   \includegraphics[width=0.48\linewidth]{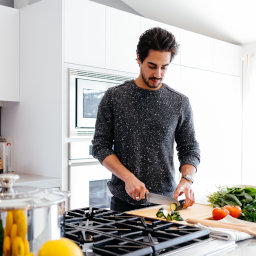}
   \includegraphics[width=0.48\linewidth]{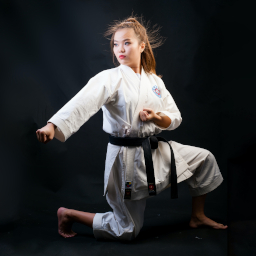}\\
   \includegraphics[width=0.48\linewidth]{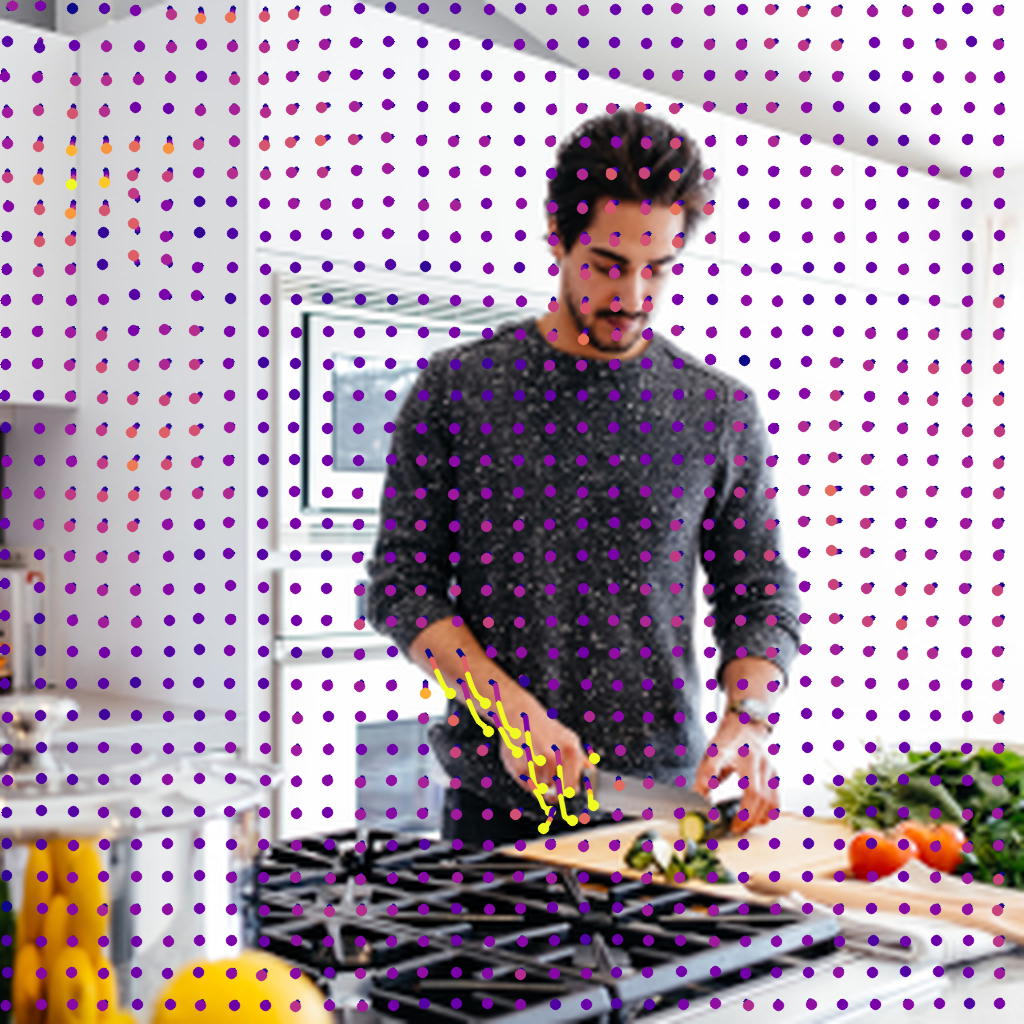}
   \includegraphics[width=0.48\linewidth]{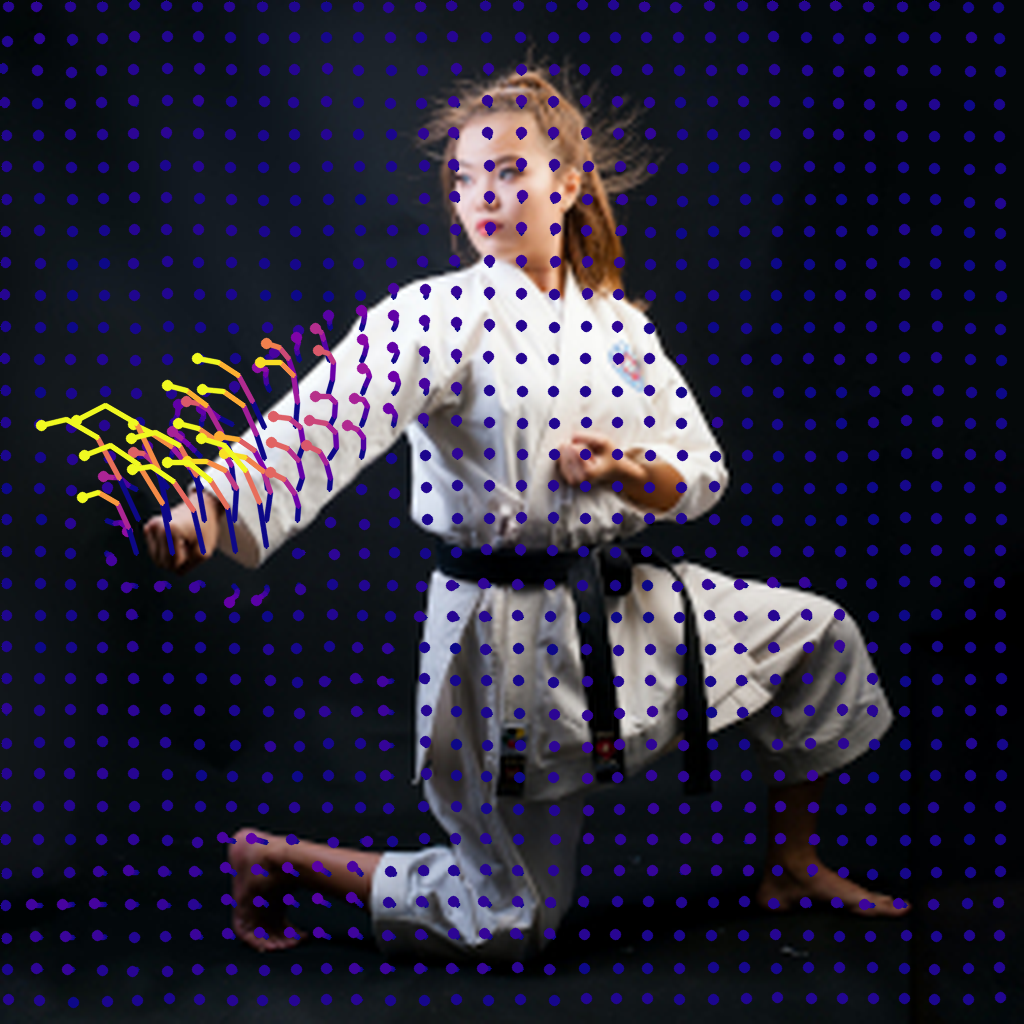}\\
   \includegraphics[width=0.48\linewidth]{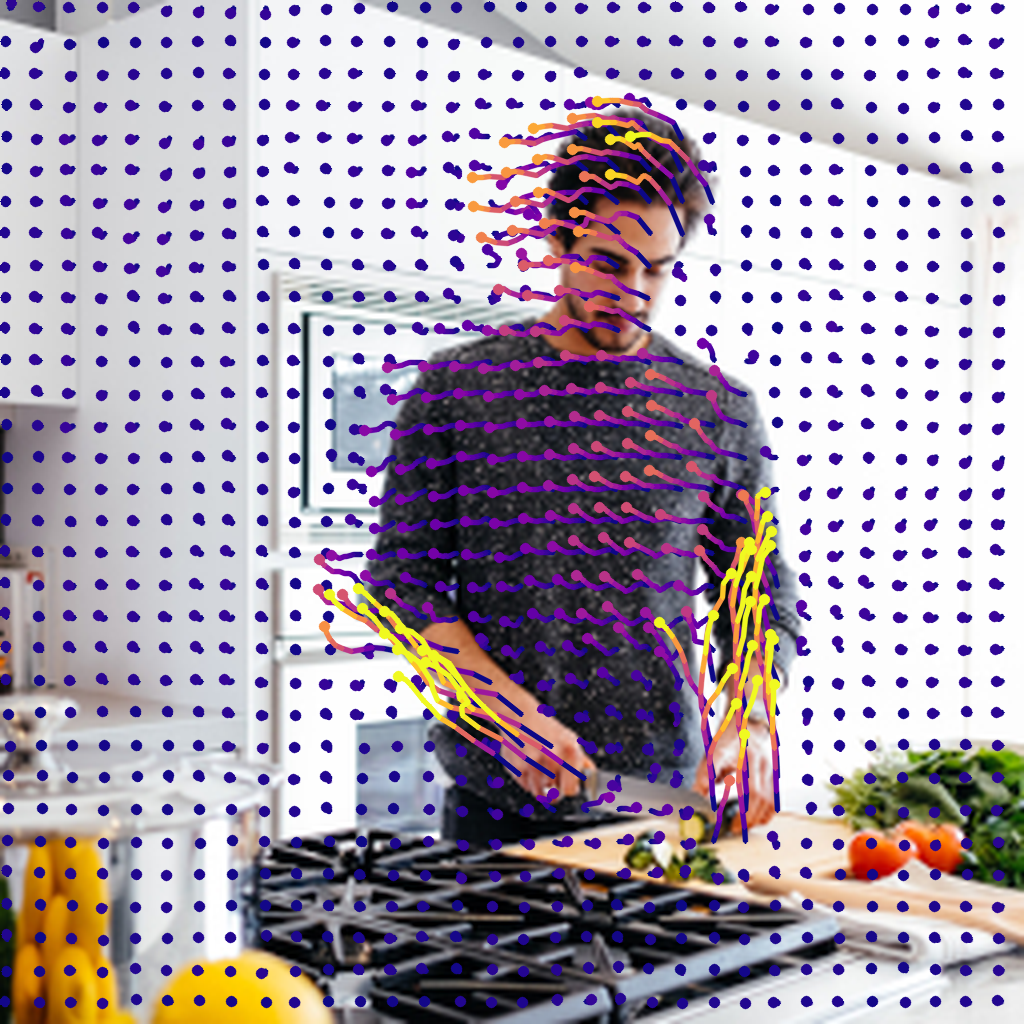}
   \includegraphics[width=0.48\linewidth]{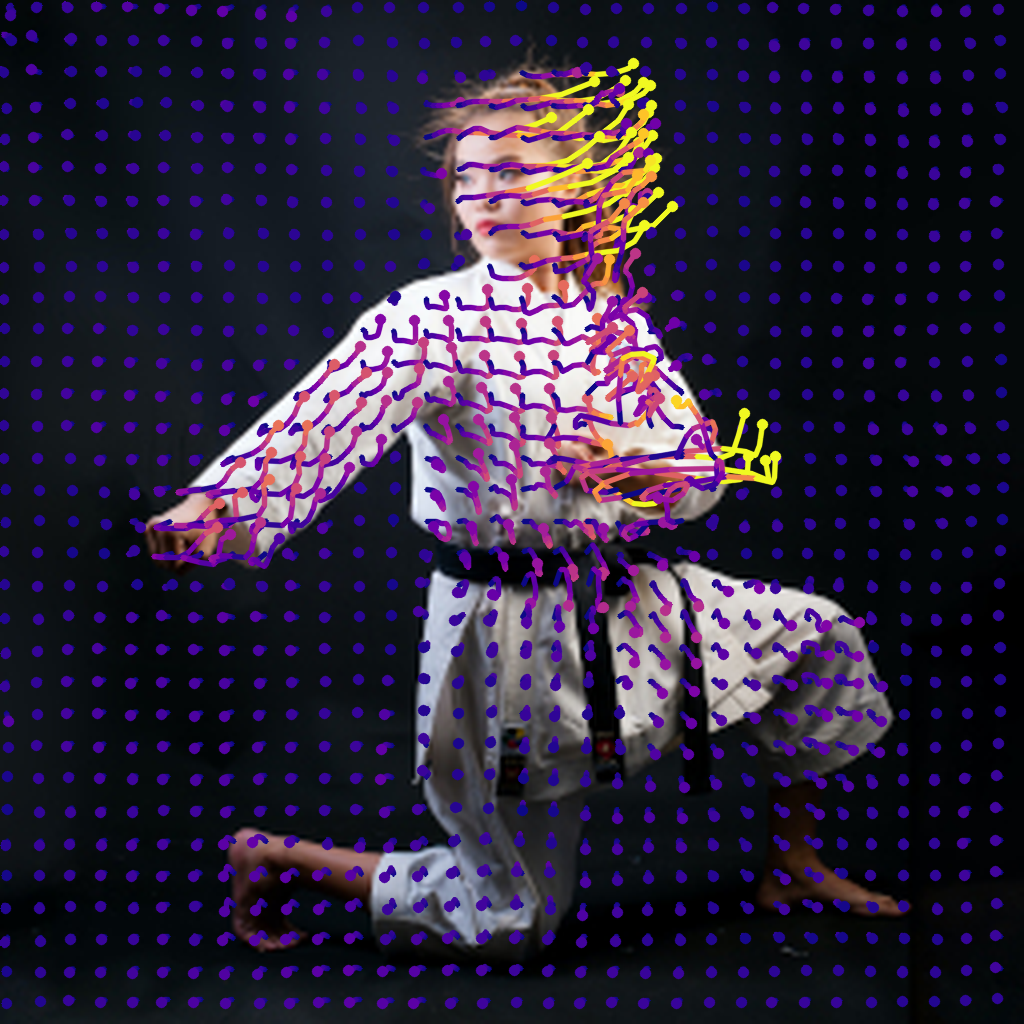}
\end{center}
   \caption{\textbf{Animating still frames.} Top row shows the input images. Each following row shows a visualization of a single sample from our trajectory model: the dark purple end is the starting point of the trajectory, while the colors get brighter and yellower with distance.  In the left image, one sample shows the hand moving forward in a cutting motion, with a slight incline of the head; in the other, the entire person moves to the left.  On the right, the first sample moves one arm in a circle, while the second shows the head turning and the other arm being raised.}
\label{fig:animate}
\end{figure}

Fig.~\ref{fig:animate} shows the trajectories produced by our trajectory prediction model on selected images.  
Our model can make multiple different physically-plausible predictions for motion for each input image, suggesting that it can recognize people and do rudimentary human pose estimation without supervision.  Note that this demonstrates two levels of transfer: TAPIR was never trained on humans, while the trajectory-prediction model was trained on video rather than still images like these.  It is difficult to judge these motions from a static figure, so we encourage the reader to view videos generated with our pipeline in the supplementary.

\section{Conclusion}
In this paper, we introduce TAPIR, a novel model for Tracking Any Point (TAP) that adopts a two-stage approach, comprising a matching stage and a refinement stage.  It demonstrates a substantial improvement over the state-of-the-art, providing stable predictions and occlusion robustness, and scales to high-resolution videos. We also show a proof-of-concept of animating still frames, which we hope can serve as a foundation for future graphics research. %
TAPIR still has some limitations, such as difficulty in precisely perceiving the boundary between background and foreground, and it may still struggle with large appearance changes. Nevertheless, our findings indicate a promising future for TAP in computer vision.

\paragraph{Acknowledgements} We would like to thank Lucas Smaira and Larisa Markeeva for help with datasets and codebases, and Jon Scholz and Todor Davchev for advice on the needs of robotics.  We also thank Adam Harley, Viorica Patraucean and Daniel Zoran for helpful discussions, and Mateusz Malinowski for paper comments.  Finally, we thank Nando de Freitas for his support.

{\small
\bibliographystyle{ieee_fullname}
\bibliography{egbib}

\begin{thebibliography}{10}\itemsep=-1pt

\bibitem{balntas2017hpatches}
Vassileios Balntas, Karel Lenc, Andrea Vedaldi, and Krystian Mikolajczyk.
\newblock Hpatches: A benchmark and evaluation of handcrafted and learned local
  descriptors.
\newblock In {\em Proceedings of IEEE Conference on Computer Vision and Pattern
  Recognition (CVPR)}, pages 5173--5182, 2017.

\bibitem{bartlett2006automatic}
Marian~Stewart Bartlett, Gwen Littlewort, Mark~G Frank, Claudia Lainscsek,
  Ian~R Fasel, Javier~R Movellan, et~al.
\newblock Automatic recognition of facial actions in spontaneous expressions.
\newblock {\em J. Multim.}, 1(6):22--35, 2006.

\bibitem{bay2006surf}
Herbert Bay, Tinne Tuytelaars, and Luc~Van Gool.
\newblock Surf: Speeded up robust features.
\newblock In {\em Proceedings of European Conference on Computer Vision
  (ECCV)}, pages 404--417. Springer, 2006.

\bibitem{brock2021high}
Andy Brock, Soham De, Samuel~L Smith, and Karen Simonyan.
\newblock High-performance large-scale image recognition without normalization.
\newblock In {\em Proceedings of International Conference on Machine Learning
  (ICML)}, pages 1059--1071. PMLR, 2021.

\bibitem{brox2009large}
Thomas Brox, Christoph Bregler, and Jitendra Malik.
\newblock Large displacement optical flow.
\newblock In {\em Proceedings of IEEE Conference on Computer Vision and Pattern
  Recognition (CVPR)}, pages 41--48. IEEE, 2009.

\bibitem{brox2004high}
Thomas Brox, Andr{\'e}s Bruhn, Nils Papenberg, and Joachim Weickert.
\newblock High accuracy optical flow estimation based on a theory for warping.
\newblock In {\em Proceedings of European Conference on Computer Vision
  (ECCV)}, pages 25--36. Springer, 2004.

\bibitem{butler2012naturalistic}
Daniel~J Butler, Jonas Wulff, Garrett~B Stanley, and Michael~J Black.
\newblock A naturalistic open source movie for optical flow evaluation.
\newblock In {\em Proceedings of European Conference on Computer Vision
  (ECCV)}, pages 611--625. Springer, 2012.

\bibitem{carreira2016human}
Joao Carreira, Pulkit Agrawal, Katerina Fragkiadaki, and Jitendra Malik.
\newblock Human pose estimation with iterative error feedback.
\newblock In {\em Proceedings of IEEE Conference on Computer Vision and Pattern
  Recognition (CVPR)}, pages 4733--4742, 2016.

\bibitem{carreira2017quo}
Joao Carreira and Andrew Zisserman.
\newblock Quo vadis, action recognition? a new model and the kinetics dataset.
\newblock In {\em Proceedings of IEEE Conference on Computer Vision and Pattern
  Recognition (CVPR)}, pages 6299--6308, 2017.

\bibitem{cheng2021equivariant}
Zezhou Cheng, Jong-Chyi Su, and Subhransu Maji.
\newblock On equivariant and invariant learning of object landmark
  representations.
\newblock In {\em Proceedings of IEEE International Conference on Computer
  Vision (ICCV)}, pages 9897--9906, 2021.

\bibitem{dai2017scannet}
Angela Dai, Angel~X Chang, Manolis Savva, Maciej Halber, Thomas Funkhouser, and
  Matthias Nie{\ss}ner.
\newblock Scannet: Richly-annotated 3d reconstructions of indoor scenes.
\newblock In {\em Proceedings of IEEE Conference on Computer Vision and Pattern
  Recognition (CVPR)}, pages 5828--5839, 2017.

\bibitem{detone2018superpoint}
Daniel DeTone, Tomasz Malisiewicz, and Andrew Rabinovich.
\newblock Superpoint: Self-supervised interest point detection and description.
\newblock In {\em Proceedings of IEEE Conference on Computer Vision and Pattern
  Recognition Workshops (CVPRW)}, pages 224--236, 2018.

\bibitem{doersch2022tap}
Carl Doersch, Ankush Gupta, Larisa Markeeva, Adri{\`a} Recasens, Lucas Smaira,
  Yusuf Aytar, Jo{\~a}o Carreira, Andrew Zisserman, and Yi Yang.
\newblock Tap-vid: A benchmark for tracking any point in a video.
\newblock {\em Proceedings of Neural Information Processing Systems Datasets
  and Benchmarks Track}, 2022.

\bibitem{doersch2019sim2real}
Carl Doersch and Andrew Zisserman.
\newblock Sim2real transfer learning for 3d human pose estimation: motion to
  the rescue.
\newblock {\em Proceedings of Neural Information Processing Systems (NeurIPS)},
  32, 2019.

\bibitem{dosovitskiy2015flownet}
Alexey Dosovitskiy, Philipp Fischer, Eddy Ilg, Philip Hausser, Caner Hazirbas,
  Vladimir Golkov, Patrick Van Der~Smagt, Daniel Cremers, and Thomas Brox.
\newblock Flownet: Learning optical flow with convolutional networks.
\newblock In {\em Proceedings of IEEE International Conference on Computer
  Vision (ICCV)}, pages 2758--2766, 2015.

\bibitem{geiger2012we}
Andreas Geiger, Philip Lenz, and Raquel Urtasun.
\newblock Are we ready for autonomous driving? the kitti vision benchmark
  suite.
\newblock In {\em Proceedings of IEEE Conference on Computer Vision and Pattern
  Recognition (CVPR)}, pages 3354--3361. IEEE, 2012.

\bibitem{kubric}
Klaus Greff, Francois Belletti, Lucas Beyer, Carl Doersch, Yilun Du, Daniel
  Duckworth, David~J Fleet, Dan Gnanapragasam, Florian Golemo, Charles
  Herrmann, Thomas Kipf, Abhijit Kundu, Dmitry Lagun, Issam Laradji,
  Hsueh-Ti~(Derek) Liu, Henning Meyer, Yishu Miao, Derek Nowrouzezahrai, Cengiz
  Oztireli, Etienne Pot, Noha Radwan, Daniel Rebain, Sara Sabour, Mehdi S.~M.
  Sajjadi, Matan Sela, Vincent Sitzmann, Austin Stone, Deqing Sun, Suhani Vora,
  Ziyu Wang, Tianhao Wu, Kwang~Moo Yi, Fangcheng Zhong, and Andrea
  Tagliasacchi.
\newblock Kubric: a scalable dataset generator.
\newblock In {\em Proceedings of IEEE Conference on Computer Vision and Pattern
  Recognition (CVPR)}, 2022.

\bibitem{gu2022vector}
Shuyang Gu, Dong Chen, Jianmin Bao, Fang Wen, Bo Zhang, Dongdong Chen, Lu Yuan,
  and Baining Guo.
\newblock Vector quantized diffusion model for text-to-image synthesis.
\newblock In {\em Proceedings of IEEE Conference on Computer Vision and Pattern
  Recognition (CVPR)}, pages 10696--10706, 2022.

\bibitem{densepose}
R{\i}za~Alp G{\"u}ler, Natalia Neverova, and Iasonas Kokkinos.
\newblock Densepose: Dense human pose estimation in the wild.
\newblock In {\em Proceedings of IEEE Conference on Computer Vision and Pattern
  Recognition (CVPR)}, pages 7297--7306, 2018.

\bibitem{harley2022particle}
Adam~W Harley, Zhaoyuan Fang, and Katerina Fragkiadaki.
\newblock Particle video revisited: Tracking through occlusions using point
  trajectories.
\newblock In {\em Proceedings of European Conference on Computer Vision
  (ECCV)}, pages 59--75. Springer, 2022.

\bibitem{hartley2003multiple}
Richard Hartley and Andrew Zisserman.
\newblock {\em Multiple view geometry in computer vision}.
\newblock Cambridge university press, 2003.

\bibitem{ho2020denoising}
Jonathan Ho, Ajay Jain, and Pieter Abbeel.
\newblock Denoising diffusion probabilistic models.
\newblock {\em Proceedings of Neural Information Processing Systems (NeurIPS)},
  33:6840--6851, 2020.

\bibitem{ho2022video}
Jonathan Ho, Tim Salimans, Alexey Gritsenko, William Chan, Mohammad Norouzi,
  and David~J Fleet.
\newblock Video diffusion models.
\newblock {\em arXiv:2204.03458}, 2022.

\bibitem{horn1981determining}
Berthold~KP Horn and Brian~G Schunck.
\newblock Determining optical flow.
\newblock {\em Artificial Intelligence}, 17(1-3):185--203, 1981.

\bibitem{ilg2017flownet}
Eddy Ilg, Nikolaus Mayer, Tonmoy Saikia, Margret Keuper, Alexey Dosovitskiy,
  and Thomas Brox.
\newblock Flownet 2.0: Evolution of optical flow estimation with deep networks.
\newblock In {\em Proceedings of IEEE Conference on Computer Vision and Pattern
  Recognition (CVPR)}, pages 2462--2470, 2017.

\bibitem{jakab2018unsupervised}
Tomas Jakab, Ankush Gupta, Hakan Bilen, and Andrea Vedaldi.
\newblock Unsupervised learning of object landmarks through conditional image
  generation.
\newblock {\em Proceedings of Neural Information Processing Systems (NeurIPS)},
  31, 2018.

\bibitem{jakab2020self}
Tomas Jakab, Ankush Gupta, Hakan Bilen, and Andrea Vedaldi.
\newblock Self-supervised learning of interpretable keypoints from unlabelled
  videos.
\newblock In {\em Proceedings of IEEE Conference on Computer Vision and Pattern
  Recognition (CVPR)}, pages 8787--8797, 2020.

\bibitem{jhuang2013towards}
Hueihan Jhuang, Juergen Gall, Silvia Zuffi, Cordelia Schmid, and Michael~J
  Black.
\newblock Towards understanding action recognition.
\newblock In {\em Proceedings of IEEE International Conference on Computer
  Vision (ICCV)}, pages 3192--3199, 2013.

\bibitem{jiang2021cotr}
Wei Jiang, Eduard Trulls, Jan Hosang, Andrea Tagliasacchi, and Kwang~Moo Yi.
\newblock Cotr: Correspondence transformer for matching across images.
\newblock In {\em Proceedings of IEEE International Conference on Computer
  Vision (ICCV)}, pages 6207--6217, 2021.

\bibitem{lee2017desire}
Namhoon Lee, Wongun Choi, Paul Vernaza, Christopher~B Choy, Philip~HS Torr, and
  Manmohan Chandraker.
\newblock Desire: Distant future prediction in dynamic scenes with interacting
  agents.
\newblock In {\em Proceedings of IEEE Conference on Computer Vision and Pattern
  Recognition (CVPR)}, pages 336--345, 2017.

\bibitem{li2019learning}
Zhengqi Li, Tali Dekel, Forrester Cole, Richard Tucker, Noah Snavely, Ce Liu,
  and William~T Freeman.
\newblock Learning the depths of moving people by watching frozen people.
\newblock In {\em Proceedings of IEEE Conference on Computer Vision and Pattern
  Recognition (CVPR)}, pages 4521--4530, 2019.

\bibitem{li2018megadepth}
Zhengqi Li and Noah Snavely.
\newblock Megadepth: Learning single-view depth prediction from internet
  photos.
\newblock In {\em Proceedings of IEEE Conference on Computer Vision and Pattern
  Recognition (CVPR)}, pages 2041--2050, 2018.

\bibitem{lin2020tsm}
Ji Lin, Chuang Gan, Kuan Wang, and Song Han.
\newblock Tsm: Temporal shift module for efficient and scalable video
  understanding on edge devices.
\newblock {\em IEEE Transactions on Pattern Analysis and Machine Intelligence},
  2020.

\bibitem{lipson2022coupled}
Lahav Lipson, Zachary Teed, Ankit Goyal, and Jia Deng.
\newblock Coupled iterative refinement for 6d multi-object pose estimation.
\newblock In {\em Proceedings of IEEE Conference on Computer Vision and Pattern
  Recognition (CVPR)}, pages 6728--6737, 2022.

\bibitem{liu2010sift}
Ce Liu, Jenny Yuen, and Antonio Torralba.
\newblock Sift flow: Dense correspondence across scenes and its applications.
\newblock {\em IEEE Transactions on Pattern Analysis and Machine Intelligence},
  33(5):978--994, 2010.

\bibitem{lowe1999object}
David~G Lowe.
\newblock Object recognition from local scale-invariant features.
\newblock In {\em Proceedings of IEEE International Conference on Computer
  Vision (ICCV)}, volume~2, pages 1150--1157. Ieee, 1999.

\bibitem{lowe2004distinctive}
David~G Lowe.
\newblock Distinctive image features from scale-invariant keypoints.
\newblock {\em International Journal of Computer Vision}, 60(2):91--110, 2004.

\bibitem{lucas1981iterative}
Bruce~D Lucas and Takeo Kanade.
\newblock An iterative image registration technique with an application to
  stereo vision.
\newblock In {\em Proceedings of International Joint Conference on Artificial
  Intelligence (IJCAI)}, pages 674--679, 1981.

\bibitem{manuelli2020keypoints}
Lucas Manuelli, Yunzhu Li, Pete Florence, and Russ Tedrake.
\newblock Keypoints into the future: Self-supervised correspondence in
  model-based reinforcement learning.
\newblock {\em arXiv preprint arXiv:2009.05085}, 2020.

\bibitem{marchetti2020mantra}
Francesco Marchetti, Federico Becattini, Lorenzo Seidenari, and Alberto~Del
  Bimbo.
\newblock Mantra: Memory augmented networks for multiple trajectory prediction.
\newblock In {\em Proceedings of the IEEE/CVF conference on computer vision and
  pattern recognition}, pages 7143--7152, 2020.

\bibitem{marr1979computational}
David Marr and Tomaso Poggio.
\newblock A computational theory of human stereo vision.
\newblock {\em Proceedings of the Royal Society of London. Series B. Biological
  Sciences}, 204(1156):301--328, 1979.

\bibitem{mayer2016large}
Nikolaus Mayer, Eddy Ilg, Philip Hausser, Philipp Fischer, Daniel Cremers,
  Alexey Dosovitskiy, and Thomas Brox.
\newblock A large dataset to train convolutional networks for disparity,
  optical flow, and scene flow estimation.
\newblock In {\em Proceedings of IEEE Conference on Computer Vision and Pattern
  Recognition (CVPR)}, pages 4040--4048, 2016.

\bibitem{newcombe2011kinectfusion}
Richard~A Newcombe, Shahram Izadi, Otmar Hilliges, David Molyneaux, David Kim,
  Andrew~J Davison, Pushmeet Kohi, Jamie Shotton, Steve Hodges, and Andrew
  Fitzgibbon.
\newblock Kinectfusion: Real-time dense surface mapping and tracking.
\newblock In {\em 2011 10th IEEE international symposium on mixed and augmented
  reality}, pages 127--136. Ieee, 2011.

\bibitem{nichol2021glide}
Alex Nichol, Prafulla Dhariwal, Aditya Ramesh, Pranav Shyam, Pamela Mishkin,
  Bob McGrew, Ilya Sutskever, and Mark Chen.
\newblock Glide: Towards photorealistic image generation and editing with
  text-guided diffusion models.
\newblock {\em arXiv preprint arXiv:2112.10741}, 2021.

\bibitem{ono2018lf}
Yuki Ono, Eduard Trulls, Pascal Fua, and Kwang~Moo Yi.
\newblock Lf-net: Learning local features from images.
\newblock {\em Proceedings of Neural Information Processing Systems (NeurIPS)},
  31, 2018.

\bibitem{perez2018film}
Ethan Perez, Florian Strub, Harm De~Vries, Vincent Dumoulin, and Aaron
  Courville.
\newblock Film: Visual reasoning with a general conditioning layer.
\newblock In {\em Proceedings of AAAI Conference on Artificial Intelligence},
  volume~32, 2018.

\bibitem{pont20172017}
Jordi Pont-Tuset, Federico Perazzi, Sergi Caelles, Pablo Arbel{\'a}ez, Alex
  Sorkine-Hornung, and Luc Van~Gool.
\newblock The 2017 davis challenge on video object segmentation.
\newblock {\em arXiv preprint arXiv:1704.00675}, 2017.

\bibitem{ramanan2005strike}
Deva Ramanan, David~A Forsyth, and Andrew Zisserman.
\newblock Strike a pose: Tracking people by finding stylized poses.
\newblock In {\em Proceedings of IEEE Conference on Computer Vision and Pattern
  Recognition (CVPR)}, volume~1, pages 271--278. IEEE, 2005.

\bibitem{ramesh2022hierarchical}
Aditya Ramesh, Prafulla Dhariwal, Alex Nichol, Casey Chu, and Mark Chen.
\newblock Hierarchical text-conditional image generation with clip latents.
\newblock {\em arXiv preprint arXiv:2204.06125}, 2022.

\bibitem{ranjan2017optical}
Anurag Ranjan and Michael~J Black.
\newblock Optical flow estimation using a spatial pyramid network.
\newblock In {\em Proceedings of IEEE Conference on Computer Vision and Pattern
  Recognition (CVPR)}, pages 4161--4170, 2017.

\bibitem{rong2021monocular}
Yu Rong, Jingbo Wang, Ziwei Liu, and Chen~Change Loy.
\newblock Monocular 3d reconstruction of interacting hands via collision-aware
  factorized refinements.
\newblock In {\em 2021 International Conference on 3D Vision (3DV)}, pages
  432--441. IEEE, 2021.

\bibitem{Rubinstein2012}
Michael Rubinstein, Ce Liu, and William Freeman.
\newblock Towards longer long-range motion trajectories.
\newblock In {\em Proceedings of British Machine Vision Conference}, 2012.

\bibitem{saharia2022photorealistic}
Chitwan Saharia, William Chan, Saurabh Saxena, Lala Li, Jay Whang, Emily
  Denton, Seyed Kamyar~Seyed Ghasemipour, Burcu~Karagol Ayan, S~Sara Mahdavi,
  Rapha~Gontijo Lopes, et~al.
\newblock Photorealistic text-to-image diffusion models with deep language
  understanding.
\newblock {\em arXiv preprint arXiv:2205.11487}, 2022.

\bibitem{sand2008particle}
Peter Sand and Seth Teller.
\newblock Particle video: Long-range motion estimation using point
  trajectories.
\newblock {\em International Journal of Computer Vision}, 80(1):72--91, 2008.

\bibitem{schonberger2016structure}
Johannes~L Schonberger and Jan-Michael Frahm.
\newblock Structure-from-motion revisited.
\newblock In {\em Proceedings of IEEE Conference on Computer Vision and Pattern
  Recognition (CVPR)}, pages 4104--4113, 2016.

\bibitem{schops2017multi}
Thomas Schops, Johannes~L Schonberger, Silvano Galliani, Torsten Sattler,
  Konrad Schindler, Marc Pollefeys, and Andreas Geiger.
\newblock A multi-view stereo benchmark with high-resolution images and
  multi-camera videos.
\newblock In {\em Proceedings of IEEE Conference on Computer Vision and Pattern
  Recognition (CVPR)}, pages 3260--3269, 2017.

\bibitem{sethi1987finding}
Ishwar~K Sethi and Ramesh Jain.
\newblock Finding trajectories of feature points in a monocular image sequence.
\newblock {\em IEEE Transactions on Pattern Analysis and Machine Intelligence},
  (1):56--73, 1987.

\bibitem{shen2015first}
Jie Shen, Stefanos Zafeiriou, Grigoris~G Chrysos, Jean Kossaifi, Georgios
  Tzimiropoulos, and Maja Pantic.
\newblock The first facial landmark tracking in-the-wild challenge: Benchmark
  and results.
\newblock In {\em Proceedings of the IEEE International Conference on Computer
  Vision (ICCV) Workshops}, pages 50--58, 2015.

\bibitem{shrivastava2017learning}
Ashish Shrivastava, Tomas Pfister, Oncel Tuzel, Joshua Susskind, Wenda Wang,
  and Russell Webb.
\newblock Learning from simulated and unsupervised images through adversarial
  training.
\newblock In {\em Proceedings of IEEE Conference on Computer Vision and Pattern
  Recognition (CVPR)}, pages 2107--2116, 2017.

\bibitem{singer2022make}
Uriel Singer, Adam Polyak, Thomas Hayes, Xi Yin, Jie An, Songyang Zhang, Qiyuan
  Hu, Harry Yang, Oron Ashual, Oran Gafni, et~al.
\newblock Make-a-video: Text-to-video generation without text-video data.
\newblock {\em arXiv preprint arXiv:2209.14792}, 2022.

\bibitem{Sivic2006}
Josef Sivic, Frederik Schaffalitzky, and Andrew Zisserman.
\newblock Object level grouping for video shots.
\newblock In {\em Proceedings of European Conference on Computer Vision
  (ECCV)}, 2004.

\bibitem{song2020score}
Yang Song, Jascha Sohl-Dickstein, Diederik~P Kingma, Abhishek Kumar, Stefano
  Ermon, and Ben Poole.
\newblock Score-based generative modeling through stochastic differential
  equations.
\newblock {\em Proceedings of International Conference on Learning
  Representations (ICLR)}, 2021.

\bibitem{su2022zebrapose}
Yongzhi Su, Mahdi Saleh, Torben Fetzer, Jason Rambach, Nassir Navab, Benjamin
  Busam, Didier Stricker, and Federico Tombari.
\newblock Zebrapose: Coarse to fine surface encoding for 6dof object pose
  estimation.
\newblock In {\em Proceedings of IEEE Conference on Computer Vision and Pattern
  Recognition (CVPR)}, pages 6738--6748, 2022.

\bibitem{sun2021autoflow}
Deqing Sun, Daniel Vlasic, Charles Herrmann, Varun Jampani, Michael Krainin,
  Huiwen Chang, Ramin Zabih, William~T Freeman, and Ce Liu.
\newblock Autoflow: Learning a better training set for optical flow.
\newblock In {\em Proceedings of IEEE Conference on Computer Vision and Pattern
  Recognition (CVPR)}, pages 10093--10102, 2021.

\bibitem{sun2018pwc}
Deqing Sun, Xiaodong Yang, Ming-Yu Liu, and Jan Kautz.
\newblock Pwc-net: Cnns for optical flow using pyramid, warping, and cost
  volume.
\newblock In {\em Proceedings of IEEE Conference on Computer Vision and Pattern
  Recognition (CVPR)}, pages 8934--8943, 2018.

\bibitem{sun2014deep}
Yi Sun, Yuheng Chen, Xiaogang Wang, and Xiaoou Tang.
\newblock Deep learning face representation by joint
  identification-verification.
\newblock {\em Proceedings of Neural Information Processing Systems (NeurIPS)},
  27, 2014.

\bibitem{teed2020raft}
Zachary Teed and Jia Deng.
\newblock Raft: Recurrent all-pairs field transforms for optical flow.
\newblock In {\em Proceedings of European Conference on Computer Vision
  (ECCV)}, pages 402--419. Springer, 2020.

\bibitem{thewlis2017unsupervised}
James Thewlis, Hakan Bilen, and Andrea Vedaldi.
\newblock Unsupervised learning of object landmarks by factorized spatial
  embeddings.
\newblock In {\em Proceedings of IEEE International Conference on Computer
  Vision (ICCV)}, pages 5916--5925, 2017.

\bibitem{tompson2014real}
Jonathan Tompson, Murphy Stein, Yann Lecun, and Ken Perlin.
\newblock Real-time continuous pose recovery of human hands using convolutional
  networks.
\newblock {\em ACM Transactions on Graphics (ToG)}, 33(5):1--10, 2014.

\bibitem{torr1999feature}
Philip~HS Torr and Andrew Zisserman.
\newblock Feature based methods for structure and motion estimation.
\newblock In {\em International workshop on vision algorithms}, pages 278--294.
  Springer, 1999.

\bibitem{vijayanarasimhan2017sfm}
Sudheendra Vijayanarasimhan, Susanna Ricco, Cordelia Schmid, Rahul Sukthankar,
  and Katerina Fragkiadaki.
\newblock Sfm-net: Learning of structure and motion from video.
\newblock {\em arXiv preprint arXiv:1704.07804}, 2017.

\bibitem{von2018recovering}
Timo von Marcard, Roberto Henschel, Michael~J Black, Bodo Rosenhahn, and Gerard
  Pons-Moll.
\newblock Recovering accurate 3d human pose in the wild using imus and a moving
  camera.
\newblock In {\em Proceedings of European Conference on Computer Vision
  (ECCV)}, pages 601--617, 2018.

\bibitem{wang2013action}
Heng Wang and Cordelia Schmid.
\newblock Action recognition with improved trajectories.
\newblock In {\em Proceedings of IEEE International Conference on Computer
  Vision (ICCV)}, pages 3551--3558, 2013.

\bibitem{wang2019prnet}
Yue Wang and Justin~M Solomon.
\newblock Prnet: Self-supervised learning for partial-to-partial registration.
\newblock {\em Proceedings of Neural Information Processing Systems (NeurIPS)},
  32, 2019.

\bibitem{xiong2013supervised}
Xuehan Xiong and Fernando De~la Torre.
\newblock Supervised descent method and its applications to face alignment.
\newblock In {\em Proceedings of IEEE Conference on Computer Vision and Pattern
  Recognition (CVPR)}, pages 532--539, 2013.

\bibitem{xu2022gmflow}
Haofei Xu, Jing Zhang, Jianfei Cai, Hamid Rezatofighi, and Dacheng Tao.
\newblock Gmflow: Learning optical flow via global matching.
\newblock In {\em Proceedings of IEEE Conference on Computer Vision and Pattern
  Recognition (CVPR)}, pages 8121--8130, 2022.

\bibitem{xu2021rethinking}
Jiarui Xu and Xiaolong Wang.
\newblock Rethinking self-supervised correspondence learning: A video
  frame-level similarity perspective.
\newblock In {\em Proceedings of IEEE International Conference on Computer
  Vision (ICCV)}, pages 10075--10085, 2021.

\bibitem{yin2019hierarchical}
Zhichao Yin, Trevor Darrell, and Fisher Yu.
\newblock Hierarchical discrete distribution decomposition for match density
  estimation.
\newblock In {\em Proceedings of IEEE Conference on Computer Vision and Pattern
  Recognition (CVPR)}, pages 6044--6053, 2019.

\bibitem{zhang2018unsupervised}
Yuting Zhang, Yijie Guo, Yixin Jin, Yijun Luo, Zhiyuan He, and Honglak Lee.
\newblock Unsupervised discovery of object landmarks as structural
  representations.
\newblock In {\em Proceedings of IEEE Conference on Computer Vision and Pattern
  Recognition (CVPR)}, pages 2694--2703, 2018.

\bibitem{zhang2000flexible}
Zhengyou Zhang.
\newblock A flexible new technique for camera calibration.
\newblock {\em IEEE Transactions on Pattern Analysis and Machine Intelligence},
  22(11):1330--1334, 2000.

\bibitem{zhou2018stereo}
Tinghui Zhou, Richard Tucker, John Flynn, Graham Fyffe, and Noah Snavely.
\newblock Stereo magnification: Learning view synthesis using multiplane
  images.
\newblock {\em arXiv preprint arXiv:1805.09817}, 2018.

\bibitem{zimmermann2019freihand}
Christian Zimmermann, Duygu Ceylan, Jimei Yang, Bryan Russell, Max Argus, and
  Thomas Brox.
\newblock Freihand: A dataset for markerless capture of hand pose and shape
  from single rgb images.
\newblock In {\em Proceedings of IEEE International Conference on Computer
  Vision (ICCV)}, 2019.

\bibitem{zollhofer2014real}
Michael Zollh{\"o}fer, Matthias Nie{\ss}ner, Shahram Izadi, Christoph Rehmann,
  Christopher Zach, Matthew Fisher, Chenglei Wu, Andrew Fitzgibbon, Charles
  Loop, Christian Theobalt, et~al.
\newblock Real-time non-rigid reconstruction using an rgb-d camera.
\newblock {\em ACM Transactions on Graphics (ToG)}, 33(4):1--12, 2014.

\end{thebibliography}
}

\appendix

\section{Runtime Analysis}
\label{sec:runtime}

\begin{table}[t]
\resizebox{\columnwidth}{!}{%
\begin{tabular}{l|ccc|ccc|ccc}
\toprule
\multicolumn{1}{l|}{Time (s)} & \multicolumn{3}{c|}{\# points=10} & \multicolumn{3}{c|}{\# points=20} & \multicolumn{3}{c}{\# points=50} \\ 
\multicolumn{1}{l|}{} & \multicolumn{1}{l}{TAP-Net} & \multicolumn{1}{l}{PIPs} & \multicolumn{1}{l|}{TAPIR} & \multicolumn{1}{l}{TAP-Net} & \multicolumn{1}{l}{PIPs} & \multicolumn{1}{l|}{TAPIR} & \multicolumn{1}{l}{TAP-Net} & \multicolumn{1}{l}{PIPs} & \multicolumn{1}{l}{TAPIR} \\
\midrule
\# frames = 8 & 0.02 & 1.3 & 0.08 & 0.02 & 2.4 & 0.08 & 0.02 & 6.2 & 0.09 \\
\# frames = 25 & 0.05 & 3.4 & 0.12 & 0.05 & 7.2 & 0.13 & 0.05 & 17.9 & 0.15 \\
\# frames = 50 & 0.09 & 6.9 & 0.20 & 0.09 & 14.0 & 0.21 & 0.09 & 34.5 & 0.25 \\
\bottomrule
\end{tabular}%
}
\caption{\textbf{Computational time for model inference.} We conducted a comparison of computational time (in seconds) for model inference on the DAVIS video \textit{horsejump-high} with 256x256 resolution.}
\label{tab:computational_time}
\end{table}

A fast runtime is of critical importance to widespread adoption of TAP.  Dense reconstruction, for example, may require tracking densely.  

We ran TAP-Net, PIPs, and TAPIR on the DAVIS video \textit{horsejump-high}, which has a resolution of 256x256. All models are evaluated on a single V100 GPU using 5 runs on average. Query points are randomly sampled on the first frame with 10, 20, and 50 points, and the video input was truncated to 8, 25, and 50 frames. 

Table~\ref{tab:computational_time}  shows our results.  Both TAP-Net and TAPIR are capable of conducting fast inference due to their parallelization techniques; for the number of points we evaluated, the runtime was dominated by feature computation and GPU overhead, and so the runtime is independent of the number of points, and appears to be scaled better than linear across different number of frames (though parts of both algorithms do still scale linearly).  In contrast, PIPs exhibits a linear increase in computational time with respect to the number of points and frames processed. Overall, TAP-Net and TAPIR are more efficient in terms of computational time than PIPs, particularly for longer videos and a larger number of points.

\section{More Ablations}
\label{sec:ablation}
Although our model does not have many extra components, both TAP-Net and PIPs have a fair number of their own design decisions. %
In this section, we examine the importance of some of these design decisions quantitatively.  We consider four open questions.  First, are there simple and fast alternatives to `chaining' that allow the model to process the full video between the query point and every output frame?  Second, is within-channel computation (i.e.\ depthwise conv or the within-channel layers of the Mixer) important for overall performance, or do more standard dense convolutions work better?  Third, do we need as many pyramid levels as proposed in the PIPs model?  Fourth, do we still need the time shifting that was proposed in TAP-Net?  In this section, we address each question in turn.

\begin{table}[t]
\resizebox{\columnwidth}{!}{%
\begin{tabular}{l|llll}
\toprule
Average Jaccard (AJ) & Kinetics & DAVIS & RGB-Stacking & Kubric \\ 
\midrule
TAPIR Model & 57.2 & \textbf{61.3} & \textbf{62.7} & \textbf{84.7} \\
+ With RNN & 57.1 & 61.1 & 59.6 & 84.6 \\
+ With RNN with gating & \textbf{57.6} & 61.1 & 59.5 & 84.5 \\ 
\bottomrule
\end{tabular}%
}
\caption{\textbf{Comparison of the TAPIR model with and without an RNN.}  We see relatively little benefit for this kind of temporal integration, though we do not see a detriment either.  This suggests an area for future research.}
\label{tab:rnn_compare}
\end{table}

\subsection{Recurrent Neural Networks}
In theory, the `chaining' initialization of PIPs has an advantage which TAPIR does not reproduce: when the model has a query on frame $t_q$ and makes a prediction at a later frame $t_o$, the prediction at frame $t_o$ depends on \textit{all} frames between $t_q$ and $t_o$.  This may be advantageous if there are no occlusions between $t_q$ and $t_o$, and the point's appearance changes over time so that it becomes difficult to match based on appearance alone.  PIPs can, in these cases, use temporal continuity to avoid losing track of the target point, in a way that TAPIR cannot do easily.

Of course, it's unclear how important this is in practice: relying on temporal continuity can backfire under occlusions.  To begin to investigate this question, we made a simple addition to the model: an RNN which operates across time, which has similar properties to the `chaining,' which we apply during the TAP-Net initialization after computing the cost volume.

A convolutional recurrent neural network (Conv RNN) is a classic architecture for spatiotemporal reasoning, and could integrate information across the entire video.  However, if given direct access to the features, it could overfit to the object categories; furthermore, if given too much latent state, it could easily memorize the stereotypical motion patterns of the Kubric dataset.  Global scene motion, on the other hand, can give cues for the motion of points that are weakly textured or occluded.  Our RNN architecture aims to capture both temporal continuity and global motion while avoiding overfitting.

The first step is to compute global motion statistics.  Starting with the original feature map $F\in{\mathbb{R}^{T\times H//8\times W//8\times D}}$, we compute local correlations $D\in{\mathbb{R}^{T\times H\times W\times 9}}$ for each feature in frame $t$ with nearby features in frame $t+1$:

\begin{align}
\begin{split}
S_{t,x,y}=\{F_{t,x,y}\cdot F_{t-1,i,j}| & x-1\leq i \leq x+1; \\
                                             & y-1\leq j \leq y+1\}
\end{split}
\end{align}

Note that, for the first frame, we wrap $t-1$ to the final frame.  We pass $S$ into two Conv+ReLU layers (256 channels) before global average pooling and projecting the result to $P_{t}\in\mathbb{R}^{32}$ dimensions.  Note that this computation does not depend on the query point, and thus can be computed just once.  The resulting 32-dimensional vector can be seen as a summary of the motion statistics, which will be used as a gating mechanism for propagating matches.

The RNN is applied separately for every query point.  The state $R_t\in{\mathbb{R}^{H\times W\times 1}}$ of the RNN itself is a simple spatial softmax, except for the first frame which is initialized to 0.  The cost volume $C_{t}$ at time $t$ is first passed through a conv layer (which increases the dimensionality to $32$ channels), and then concatenated with the RNN state $R_{t-1}$ from time $t-1$.   This is passed through a Conv layer with 32 channels, which is then multiplied by the (spatially broadcasted) motion statistics $P^{t}$.  Then a final Conv (1 channel) followed by a Softmax computes $R_{t}$.  The final output of the RNN is the simple concatenation of the state $R$ across all times. The initial track estimate for each frame is a soft argmax of this heatmap, as in TAPIR.  We do not modify the occlusion or uncertainty estimate part of TAPIR.

Table~\ref{tab:rnn_compare} shows our results.  We also show a simplified version, where we drop the `gating', and simply compute $R_{t}$ from $R_{t-1}$ concatenated with $C_{t}$, two Conv layers, and a softmax.  Unfortunately, the performance improvement is negligible: with gating we see just a 0.4\% improvement on Kinetics and a 0.2\% loss on DAVIS.  We were not able to find any RNN which improves over this, and informal experiments adding more powerful features to the RNN harmed performance.  While the recurrent processing makes intuitive sense, actually using it in a model appears to be non-trivial, and is an interesting area for future research.

\begin{table}[t]
\resizebox{\columnwidth}{!}{%
\begin{tabular}{l|cccc}
\toprule
Average Jaccard (AJ) & Kinetics & DAVIS & RGB-Stacking & Kubric \\ 
\midrule
MLP Mixer & 54.9 & 53.8 & 61.9 & 79.7 \\
Conv1D & 56.9 & 60.9 & 61.3 & 84.6 \\
Depthwise Conv & \textbf{57.2} & \textbf{61.3} & \textbf{62.7} & \textbf{84.7} \\ 
\bottomrule
\end{tabular}%
}
\caption{\textbf{Comparison between the layer kernel type in iterative updates.} We find that depthwise conv works slightly better than a dense 1D convolution, even though the latter has more parameters.}
\label{tab:conv_kernel_compare}
\end{table}

\subsection{Depthwise versus Dense Convolution}

PIPs uses an MLP-Mixer for its iterative updates to the tracks, and the within-channel layers of the mixer inspired the depthwise conv layers of TAPIR.  How critical is this design choice?  It's desirable as it saves some compute, but at the same time, it reduces the expressive power relative to a dense 1D convolution.  To answer this question, we replaced all depthwise conv layers in TAPIR with dense 1D convolutions of the same shape, a decision which almost quadruples the number of parameters in our refinement network.  The results in table~\ref{tab:conv_kernel_compare}, however, show that this is actually slightly harmful to performance, although the differences are negligible.  This suggests that, despite adding more expressivity to the network, in practice it might result in overfitting to the domain, possibly by allowing the model to memorize motion patterns that it sees in Kubric.  Using models with more parameters is an interesting direction for future work.

\begin{table}[t]
\resizebox{\columnwidth}{!}{%
\begin{tabular}{l|cccc}
\toprule
Average Jaccard (AJ) & Kinetics & DAVIS & RGB-Stacking & Kubric \\ 
\midrule
pyramid\_level=5 (TAPIR) & 57.2 & 61.3 & 62.7 & 84.7 \\ 
pyramid\_level=4 & 57.5 & 61.3 & 61.6 & \textbf{84.9} \\
pyramid\_level=3 & \textbf{57.9} & \textbf{61.5} & \textbf{63.0} & 84.8 \\
pyramid\_level=2 & 57.2 & 61.0 & 61.2 & 84.4 \\
\bottomrule
\end{tabular}%
}
\caption{\textbf{Comparison on number of feature pyramid levels.} We find that the number of pyramid levels makes relatively little difference in performance; in fact, 3 pyramid levels seems to be all that's required, and even 2 levels gives competitive performance.}
\label{tab:pyramid_level_compare}
\end{table}

\subsection{Number of Pyramid Layers}
\label{sec:pyramid_layers}
In PIPs, a key motivation for using a large number of pyramid layers is to provide more spatial context.  This is important if the initialization is poor: in such cases, the model should consider the query point similarity to other points over a wide spatial area.  TAPIR reproduces this decision, extracting a full feature pyramid for the video by max pooling, and then comparing the query feature to local neighborhoods in each layer of the pyramid.  However, TAPIR's initialization involves comparing the query to all other features in the entire frame.  Given this relatively stronger initialization, is the full pyramid necessary?  

To explore this, we applied the same TAPIR architecture, but successively removed the highest levels of the pyramid, and table~\ref{tab:pyramid_level_compare} gives the results.  We see that fewer pyramid levels than were used in the full TAPIR model are sufficient.  In fact, 2 pyramid levels saves computation while providing competitive performance.  

\begin{table}[t]
\resizebox{\columnwidth}{!}{%
\begin{tabular}{l|llll}
\toprule
Average Jaccard (AJ) & Kinetics & DAVIS & RGB-Stacking & Kubric \\ 
\midrule
TAPIR Model & \textbf{57.2} & \textbf{61.3} & \textbf{62.7} & \textbf{84.7} \\
- No TSM  & 57.1 & 61.0 & 59.4 & 84.3 \\ 
\bottomrule
\end{tabular}%
}
\caption{Comparing the TAPIR model with a version without the TSM (temporal shift module).}
\label{tab:tsm_shift_compare}
\end{table}

\subsection{Time-Shifting}
\label{sec:time_shifting}
One slight complexity that TAPIR inherits from TAP-Net is its use of a TSM-ResNet~\cite{lin2020tsm} rather than a simple ResNet.  TAP-Net's TSM-ResNet is actually modified from the original version, in that only the lowest layers of the network use the temporal shifting.  Arguably, the reason for this choice is to perform some minor temporal aggregation, but it comes with the disadvantage that it makes the model more difficult to apply in online settings, as the features for a given frame cannot be computed without seeing future frames.  However, our model uses a refinement step that's aware of time.  This raises the question: how important is the TSM module?

To answer this question, we replaced the TSM-ResNet with a regular ResNet--i.e., we simply removed the time shifting from its earliest layers, and kept all other architectural details the same.  In Table~\ref{tab:tsm_shift_compare}, we see that this actually makes little difference for TAPIR on real data, losing a negligible 0.1\% performance on Kinetics and a similarly negligible 0.3\% on DAVIS.  Only for RGB-Stacking does it seem to make a difference.  One possible explanation is that the model struggles to segment the low-texture RGB-Stacking objects, so the model uses motion cues to do this.  Regardless, it seems that for real data, the time-shifting layers can be safely removed.

\section{Implementation Details}
\label{sec:impl}
In this section, we provide implementation details to enable reproducibility, first for TAPIR, then for our new synthetic dataset.

\subsection{TAPIR}
\label{sec:tapir_details}
The TSM-ResNet model used in the bulk of experiments (but not the open-source model) follows the one in TAP-Net: i.e., it has time-shifting in the first two blocks, and replaces the striding in later blocks with dilated convolutions to achieve a stride-8 convolutional feature grid.  We use the output features of unit 2 (stride 8) and unit 0 (stride 4), and normalize both across channels before further processing.

We use the unit 2 features to compute the cost volume, using bilinear interpolation to obtain a query feature, before computing dot products with all other features in the video.  The TAP-Net-style post-processing first apply an embedding convolution (16 channels followed by ReLU) to obtain an embedding $e$, followed by a convolution to a single channel, to which we apply a spatial softmax followed by a soft argmax operation to obtain a position estimate $p^{0}_{t}$ for time $t$.  To predict occlusion, we apply a strided convolution to the embedding $e$ (32 channels) followed by a ReLU and then a spatial global average pool.  Then we apply a MLP (256 units, ReLU, and finally a projection to 2 units) to produce the logits $o^{0}_{t}$ and $u^{0}_{t}$ for occlusion and uncertainty estimate, respectively. 

The above serves as the initialization for the refinement, along with the raw features.  Each iteration $i$ produces an update $(\Delta p^i_t, \Delta o^i_t, \Delta u^i_t, \Delta F_{q,t,i})$.  To construct the inputs for the depthwise convolutional network, $o^i_t$  and $u^i_t$ are passed in the form of raw logits; $F_{q,t,i}$ is passed in unmodified.  Note that, for versions of the network with a higher resolution, $F_{q,t,i}$ comprises both the high-resolution (64-channel) feature as well as the low-resolution (256-dimensional) feature, which are concatenated along the channel axis.  The dot products that make up the score maps are passed in spatially unraveled.  The final input for $p^{i}_t$ needs a slight modification.  For $p^i_t$, PIPs subtracts the initial estimate for each chunk before processing with an MLP mixer.  We cannot follow this exactly because our model has no chunks; therefore, we instead subtract the temporal mean $x$ and $y$ value across the entire segment, ignoring occlusion.  This choice means that, like it is for PIPs, the input to the depthwise-convolutional network input is roughly invariant to spatial shifts of the entire video, at least up to truncation of the score maps.  

Following PIPs, our depthwise convolutional network first increases the number of channels to $512$ per frame with a linear projection.  Then we have 12 blocks, each consisting of one $1\times 1$ convolutional residual unit, and one depthwise residual unit.  The 12 blocks are followed by a projection back down to the output dimension, which is split into $(\Delta p^i_t, \Delta o^i_t, \Delta u^i_t, \Delta F_{q,t,i})$.  Both types of residual unit consist of a single convolution which increases the per-frame dimension to $2048$, followed by a GeLU, followed by another convolution that reduces the dimension back to $512$ before adding the input.  Note that increasing the dimensionality is non-trivial within a depthwise convolutional network: in principle, each output channel in a depthwise convolution should correspond to exactly one input channel.  Therefore, we accomplish this by running \emph{four} depthwise convolutions in parallel, each on the same input.  We apply a GeLU, and then run a second depthwise convolution on each of these outputs, and sum all four output layers.  This has the effect of increasing the per-frame dimension to $2048$ before reducing it back to $512$, while ensuring that each output channel receives only input from the same input channel.  

Our losses are described in the main text; we use the same relative scaling of the Huber and BCE as in TAP-Net, and the uncertainty loss is weighted the same as the occlusion.  Like TAP-Net, we train on 256 query points per video, sampled using the defaults from Kubric.  However, we find that running refinement on all query points with a batch size of 4 tends to run out of memory.  We found it was effective to run iterative refinement on only 32 query points per batch, randomly sampled at every iteration. 

\subsection{Training Dataset}
\label{sec:training_dataset}
Our primary motivation for generating a new dataset is a particular degenerate behavior that we noticed when running the model densely: videos with panning often caused TAPIR to fail on background points that went offscreen early in the video.  Therefore, we made a minor modification to the public MOVi-E script by changing the camera rotation.  Specifically, we set the `look at' point to follow a linear trajectory near the bottom of the workspace, traveling through the center of the workspace to ensure that the camera pans but still remains mostly looking at the objects in the scene.

To accomplish this, we first sample a `start point' $a$ within a medium-sized sphere (4 unit radius; the camera center itself starts between 6 and 12 units from the workspace center).  We also constrain that the start point $a$ is above the ground plane, but close to the ground (max 1 unit height).  Then we sample a `travel through' point $b$ in a small sphere in the center of the workspace (1 unit radius), ensuring that $b$ is also above the ground plane; the final `look at' path will travel through this point. Finally, we sample an end point by extending the line from the start to the `travel through' point by as much as 50\% of the distance between them: i.e., the `end point' $c=b+\gamma*(b-a)$, where $\gamma$ is randomly sampled between 0 and 0.5.  The final `look at' point travels on a linear trajectory, either from $a$ to $c$ or from $c$ to $a$ with a 50\% probability.  We sample 100K videos to use as a training set, and keep all other parameters consistent with Kubric.

\section{Diffusion Models for Animating Still Images}

Recent pipelines using diffusion for video generation have typically used image pretraining, as well as multiple levels model `chaining', i.e., one model may be trained to produce low-resolution, low-framerate videos, another may be used to fill in gaps between frames, a third may be used to upsample, and so on.  Make-a-video~\cite{singer2022make}, for example, had four different models, and pretrained on 2.3B images.  It also uses the same `noise' vector on every image to encourage temporal coherence, although this in practice can limit the variability of background textures across the video.  For simplicity and for computational reasons, in this project we chain together just two models, and use no image pretraining.  We expect that further model chaining, pretraining, and larger video datasets would improve performance, but that is beyond the scope of the current work.  

As described in the main paper, our diffusion model consists of two components: a \emph{trajectory} model, and a \emph{video} model.  To train both, we first run TAPIR on a large database of videos, center-cropped to 256x256 and clipped to 24 frames.  We query TAPIR with a dense, $64\times 64$ grid of query points on the first frame.  This provides training data for both models.

These two models are trained independently: the trajectory model is conditioned on the first frame and is trained to reproduce the associated trajectories.  The video model is conditioned on both the first frame and the associated trajectories, and is trained to reproduce the later frames of the video.  We describe the network architecture for each in turn.

\subsubsection{Trajectory Model}
The trajectory model first processes the input image with a modified NFNet~\cite{brock2021high} F0, modified to output a high-resolution grid: all layers after the first have striding removed and, to prevent a resulting explosion in memory, the number of channels in the backbone is capped at 512.  This results in a feature at stride 8, with relatively small receptive fields; to enable more global reasoning, we apply a multi-headed self-attention layer with 4 heads and 128 attention channels, resulting in a new feature map with 512 channels at stride 8.  The tracks, meanwhile, are at stride 4: therefore, we upsample the above feature map with a transposed convolution with 256 output channels.  To recover high-resolution information, we also apply a convolution to the NFNet's block 0 output (a stride-4 tensor with 256 channels) and add this to the output of the transposed convolution.  Thus, the final image representation $G$ is a stride 4 tensor with 256 channels.

Our overall diffusion pipeline follows DDPM~\cite{ho2020denoising} with a cosine rule.  The input array to be denoised has shape ($24 \times 64\times 64 \times 3$, where 24 is the number of frames, and 3 corresponds to x, y, and occlusion).  This encodes the $(x,y)$ positions relative to the first frame, scaled to the range $[-1,1]$ (i.e., 1 corresponds to a positive motion of 256 pixels, the max possible).  Occlusion is unfortunately a binary value, so we apply a smoothing operation to make it continuous.  Let $\tilde{o}_t$ be an occlusion indicator: i.e., 1 if the point is occluded at time $t$ and -1 otherwise.  For each point $t$, let $\hat{t}$ be the nearest time such that $\tilde{o}_{\hat{t}}\neq \tilde{o}_t$.  We compute $\bar{o}_t=\tilde{o}_t*(1-(2/3)^{|t-\hat{t}|})$.  Thus, this value decays exponentially toward the extreme values 1 and -1 as distance from a `transition' increases, but it preserves the sign of $\tilde{o}_t$, making decoding straightforward.  We use $\bar{o}_t$ as the occlusion estimate, rescaled to the range $(0,.2)$.  This rescaling encourages the model to reconstruct the motion first, and then reconstruct the occlusion value based on the reconstructed motion.

For each training example, we randomly sample a `time' value $\tau$ uniformly between 0 and 1, and we sample a noise vector $\hat{\epsilon}$ which is the same size as the input array.  The loss is then:

$$\|x_0-f_\theta(x_0*cos(\tau*\pi/2) + \hat{\epsilon}*sin(\tau*\pi/2)|G)$$

Here, $f_\theta$ is the neural network parameterized by $\theta$.  $f_{\theta}$ is a U-Net with self attention, following~\cite{song2020score}, although we use only 3 residual blocks per resolution.  To apply the conditioning, we follow the conditional group norm formulation~\cite{ho2020denoising,perez2018film}.  That is, after group norm does mean subtraction and variance normalization within each group to produce a normalized output $Z$, the layer would typically apply a scale and shift operation.  These are replaced with linear projections of the conditioning $G$.  Prior work, however, assumes that the conditioning is a single vector, i.e., $G\in R^{C}$, whereas we have a spatial tensor $G\in R^{H//4\times W//4 \times C}$.  Therefore, we first resize $G$ so that its spatial dimensions are the same size as $Z$, and then we apply two $1 \times 1$ convolution layers to create a scale and multiplier that are the same size as $Z$.

To visualize the output of this model, we use a patch-based warping.  For a given frame $t$, the trajectories at time $t$ tell us where each $4\times 4$ patch in the input image should appear.  Therefore, it is straightforward to construct a new image where each local patch is placed at its correct location, using bilinear interpolation to get subpixel accuracy.  However, in practice this will look bad if objects get larger, e.g., as they approach the camera, as there will be gaps between patches.  To deal with this, we actually warp a larger patch around each point ($8$ pixels on a side).  When multiple patches appear covering the same output pixel, we weight them inversely proportional to their distance from the track center.  This results in a smooth blend between patches that reveals motion without producing too many artifacts.  However, note that this simple method means that there tend to be artifacts around edges; e.g., points near object edges will tend to capture part of the background as well and drag it along with the object.

\subsubsection{Video Model}
The second stage of our model produces pixels given trajectories and an initial image.  At training time, we use pseudo-ground-truth trajectories from TAPIR extracted from our 24-frame videos, as well as the initial frame at $256\times 256$, and train the model to reproduce the original 24-frame clip at full $256\times 256$ resolution.  

Training the model to produce a full clip at the input resolution, however, would be prohibitive in terms of memory.  Therefore, at training time, we train our model to reconstruct a single frame at a time.  We rely mostly on the trajectory model to provide temporal coherence, at least for the image contents that are visible in the first frame.  For the rest of the pixels, we find it's beneficial to also include a small amount of temporal context from the noisy video.  

Therefore, at training time, the model must reconstruct some frame $t$.  The input is threefold.  We include two forms of `conditioning' computed by warping the input frame, first in feature space and second in image space.  Third, we input the noisy inputs for frames $t-1$ to $t+1$.  The target output is the noise residual, i.e., the noise that has been added to the image.

To compute the feature-space-warped image, we use the same image encoder that we used for the tracks model, which produces a feature grid at stride 4.  Therefore, for each position in the feature grid, we have the position where that feature appears at time $t$.  We use bilinear interpolation to place the feature at the appropriate location.  We keep track of the number of features that have been placed within any particular grid cell (specifically, the sum of bilinear interpolation weights) and normalize by this sum, unless the sum is less than 0.5, in which case we simply divide by 0.5.  This serves as a grid of conditioning features, which we use in exactly the same way that we used it in the trajectory prediction network: i.e., it is the conditioning signal for the conditional group norm.

To compute the image-space warped image, we use almost exactly the same warping algorithm that we use for our visualizations, and concatenate the result to the noisy frame.  However, we find that the model struggles to deal with the aliasing that occurs when multiple patches overlap; with just a single warp, the model cannot tell what is the contribution of each patch in turn.  Therefore, we actually warp the same patch 5 times.  The only difference between the warps is the way that the blending weight is calculated.  Let $p_{i,j,t}$ be the position of the trajectory beginning at $i,j$ in the original image at time $t$.  In the original warping, the weight for the patch carried by trajectory to any particular pixel $p^{\prime}_{t}$ (assuming $p^{\prime}_{t}$ is close enough to be within the patch) is proportional to $1/d(p_{i,j,t},p^{\prime}_{t})$, where $d$ is Euclidean distance.  The weighting that we use for the five warps is instead $1/d(p_{i,j,t}+\eta,p^{\prime}_{t})$, where $\eta\in\{(0,0),(-2,0),(0,-2),(2,0),0,2)\}$.  Therefore, the first warp is identical to the warp used in the visualization; however, the model can use the differences in intensities between the different warpings to infer the original values for different patches, and can use this information to `undo' the aliasing.

We also make a few modifications to the U-Net architecture.  Notably, we reduce the number of residual blocks per resolution from 3 to 2, and we use self attention only at the lowest-resolution layers (which are at stride 8).  We find that the memory usage is otherwise prohibitive.  We also find it's beneficial to introduce a gating layer which allows the lower-resolution pathways of the U-Net to preferentially make predictions for parts of the image that contain `holes' in the warped image; these are the parts of the image where higher-level information is most critical to `fill in' the missing data.  Recall that U-Nets combine the lower-resolution, higher-level feature tensors with `skip connection' tensors by first upsampling the lower-resolution features via a transposed strided convolution, and then concatenating it with the `skip connection' tensor.  We take the output of this concatenation and apply a projection down a single channel; we interpret this as a logit and apply a sigmoid to obtain a `gating' value between 0 and 1, with a single channel but the same spatial shape as the input tensors.  We then repeat the concatenation, but this time, multiply the skip connection by the `gating' value before concatenating.  Therefore, the model can override the low-level information with higher-level information if it decides that the lower-level input is uninformative.  

At test time, we use different noise vectors for every frame.  For every iteration of diffusion, we break the noisy video into the above, per-frame inputs and make a prediction for a single frame.  Then we concatenate all outputs and repeat the process.

We trained both models for roughly 200K iterations with a cosine learning rate schedule.  We use a batch size of 4096 for the trajectory model and a batch size of 256 for the video model, each time training on a $2\times 2 \times 2$ TPU-v3 pod.  In practice we performed relatively little tuning of this model due to its high computational cost; we suspect that further architectural tuning, as well as larger datasets, will lead to improved performance.

\end{document}